\documentclass[11pt]{article}
\newcommand{\R}{\mathbb{R}}
\usepackage[margin=1in]{geometry}
\usepackage{amsmath,amssymb,amsthm,bm}
\usepackage{booktabs,multirow,graphicx,array,makecell}
\usepackage[numbers,sort&compress]{natbib}
\usepackage{xcolor}
\usepackage{tikz}
\usepackage{adjustbox}
\usetikzlibrary{positioning,arrows.meta,calc,fit,backgrounds,decorations.pathreplacing,shapes.geometric}
\definecolor{cIn}{HTML}{4C72B0}\definecolor{cLin}{HTML}{DD8452}\definecolor{cNorm}{HTML}{55A868}
\definecolor{cFreq}{HTML}{8172B3}\definecolor{cAtt}{HTML}{C44E52}\definecolor{cPar}{HTML}{937860}
\tikzset{
  every picture/.style={font=\footnotesize},
  blk/.style={draw=#1!65!black,fill=#1!12,rounded corners=2pt,minimum height=9mm,minimum width=15mm,
              align=center,font=\footnotesize,inner sep=3pt,line width=0.6pt},
  blk/.default=gray,
  dat/.style={draw=black!70,fill=white,minimum height=9mm,minimum width=10mm,align=center,font=\footnotesize,inner sep=2pt,line width=0.6pt},
  op/.style={draw,circle,inner sep=0pt,font=\small,minimum size=5.5mm,line width=0.6pt},
  arr/.style={-{Stealth[length=2.1mm,width=1.7mm]},line width=0.65pt,draw=black!80,rounded corners=2.5pt,shorten >=0.3pt},
  darr/.style={-{Stealth[length=2.1mm,width=1.7mm]},line width=0.65pt,draw=black!80,dashed,rounded corners=2.5pt,shorten >=0.3pt},
  lbl/.style={font=\scriptsize,align=center,text=black!80},
  note/.style={font=\scriptsize,align=left,text=black!80},
  cell/.style={draw=black!50,minimum size=4mm,inner sep=0pt},
  grp/.style={draw=black!40,dashed,rounded corners=3pt,inner sep=5pt},
  sub/.style={font=\footnotesize\bfseries,anchor=west},
}

\usepackage{hyperref}
\hypersetup{colorlinks=true,linkcolor=blue!60!black,citecolor=blue!60!black,urlcolor=blue!60!black}

\newtheorem{proposition}{Proposition}
\newtheorem{corollary}{Corollary}
\newtheorem{definition}{Definition}
\theoremstyle{remark}
\newtheorem{remark}{Remark}
\newcommand{\vecr}{\operatorname{vec}}
\providecommand{\Description}[2][]{}

\title{How Many Parameters Does Forecasting Need?\\[2pt]
\large A Survey and Code-Level Audit of Lightweight Long-Term Time Series Forecasting}
\author{
Issam Ait Yahia \qquad Abdelkader El Mahdaouy \qquad Ismail Berrada\\[4pt]
{\normalsize College of Computing, Mohammed VI Polytechnic University, Ben Guerir, Morocco}\\[2pt]
{\normalsize\ttfamily \char`\{issam.aityahia, abdelkader.mahdaouy, ismail.berrada\char`\}@um6p.ma}
}
\date{}

\begin{document}
\maketitle

\begin{abstract}
Long-term time series forecasting supports decisions in energy,
transportation, weather and finance, where forecasts are produced
continuously and often on constrained hardware. Recent work has moved in two
directions. One scales Transformers and pre-trained foundation models to
millions or hundreds of millions of parameters. The other shows that models
with only hundreds to tens of thousands of parameters remain competitive on
the standard benchmarks. Existing surveys organize forecasting models by
backbone and treat efficiency as a side topic, so the reasons these small
models work, and the way their sizes are reported, remain unexamined. This
survey focuses on lightweight long-term forecasting. It provides a taxonomy
of about thirty-five models in six families, with a common notation and an
architecture diagram for each core model. It offers a structured-linear view
in which the smallest models are linear maps restricted by frequency
truncation or period folding, and it tests each restriction in closed form on
eight benchmarks. It complements the literature review with an audit of
sixteen official code repositories, which reproduces published parameter
counts and reveals reporting gaps of up to 289 times. Finally, it gives
practical guidance on model selection and reporting, and summarizes open
problems.
\end{abstract}

\section{Introduction}
\label{sec:intro}

Forecasting multivariate time series is a core task in energy and
electricity management, transportation, weather services, finance and
industrial operations \citep{benidis2023deep,emmertstreib2026forecasting}.
Long-term time series forecasting (LTSF) predicts hundreds of future steps,
typically 96 to 720, from a look-back window of similar length. Such
forecasts are recomputed continuously, for thousands of series, and
increasingly close to where the data are produced, where memory, energy and
even the storage and transmission of the data are constrained
\citep{chiarot2023compression}. Model size and inference cost are therefore
practical requirements, alongside accuracy. Parameter count is only a proxy
for these costs, and energy is rarely measured
(Sections~\ref{sec:deploy} and~\ref{sec:open}).

\begin{figure}[t]
\centering
\includegraphics[width=\linewidth]{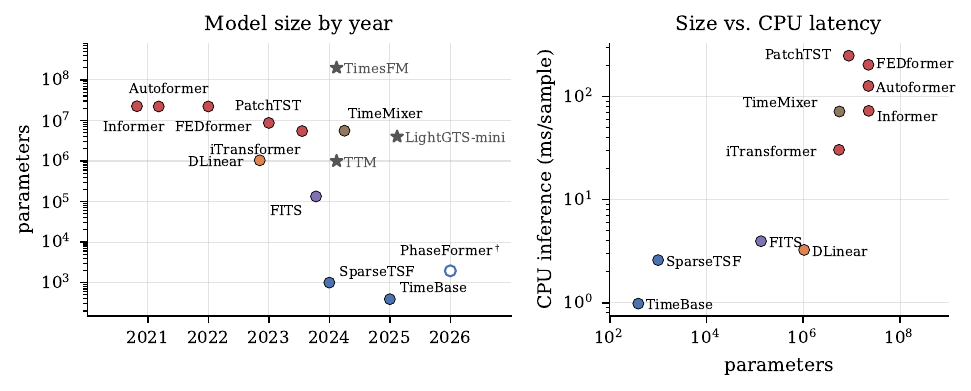}
\caption{Left: parameter counts by publication year on Electricity ($L=720$, $H=720$), as reported by \citet[Table~3]{huang2025timebase}; hollow marker: PhaseFormer counted from its code ($d=8$); stars: pre-trained models at their reported sizes \citep{das2024timesfm,ekambaram2024ttm,wang2025lightgts}. Right: parameters versus CPU inference time per sample, same source.}
\label{fig:timeline}
\Description{Two scatter plots. Left: parameter count on a log scale against publication year, falling from millions for Transformer forecasters around 2021 to hundreds for recent lightweight models, with pre-trained models shown as stars at hundreds of millions. Right: parameter count against CPU inference time per sample, both spanning several orders of magnitude.}
\end{figure}

Deep LTSF was long dominated by Transformer variants: Informer
\citep{zhou2021informer}, Autoformer \citep{wu2021autoformer}, FEDformer
\citep{zhou2022fedformer}, PatchTST \citep{nie2023patchtst}, Crossformer
\citep{zhang2023crossformer} and iTransformer \citep{liu2024itransformer}.
\citet{zeng2023dlinear} showed that a one-layer linear model is competitive.
Since then, parameter counts have fallen by three orders of magnitude:
FITS reports about $10^4$ parameters \citep{xu2024fits}, SparseTSF fewer
than $10^3$ \citep{lin2024sparsetsf}, TimeBase about $0.39$K
\citep{huang2025timebase}, and PhaseFormer about $10^3$
\citep{niu2026phaseformer}. At the same time, pre-trained foundation models
\citep{jin2026large} such as TimesFM use about 200M parameters
\citep{das2024timesfm}.
Figure~\ref{fig:timeline} shows both trends on one efficiency benchmark:
five orders of magnitude separate the largest and smallest models, and CPU
latency spans more than two orders.

\paragraph{Why a new survey.} Surveys of deep forecasting
\citep{lim2021tsf,benidis2023deep,wen2023transformers,wang2024deep,kim2025comprehensive,kong2025dtsf,emmertstreib2026forecasting}
and of its graph-based \citep{cini2025graph} and large-model
\citep{jin2026large} branches organize models by backbone (MLP, CNN, RNN,
Transformer, graph, foundation model). Companion surveys cover time series
classification \citep{foumani2024tsc}, anomaly detection
\citep{darban2024tsad}, representation learning \citep{trirat2026universal}
and diffusion models \citep{yang2026diffusion}. In the forecasting surveys
compared in Table~\ref{tab:surveys}, efficiency is a side topic: the
lightweight models of the last three years, from FITS to PhaseFormer, appear
in a few sentences or not at all. None of them asks why such small models
work, checks how their sizes are reported, or tests their structural
assumptions under one protocol. \citet{toner2024linear} analyzed the
expressivity of linear forecasters, but not the folding and cycle models
that now reach the smallest sizes. Table~\ref{tab:surveys} summarizes the difference. This survey puts parameter
efficiency at the center and asks a precise question: \emph{which structural
assumption buys which reduction, and at what cost?}

\begin{table}[t]
\centering
\caption{Comparison with related surveys and analyses. \checkmark: covered;
$\circ$: brief or partial; --: not covered.}
\label{tab:surveys}
\small
\begin{tabular}{@{}lcccccc@{}}
\toprule
Work & Venue & \makecell{Lightweight\\focus} & \makecell{Folding/\\cycle} & \makecell{Unifying\\theory} & \makecell{Code\\audit} & \makecell{Own\\experiments}\\
\midrule
\citet{lim2021tsf} & Phil.\ Trans.\ A & -- & -- & -- & -- & --\\
\citet{benidis2023deep} & ACM CSUR & -- & -- & -- & -- & --\\
\citet{cini2025graph} & ACM CSUR & -- & -- & -- & -- & \checkmark\\
\citet{jin2026large} & ACM CSUR & -- & -- & -- & -- & --\\
\citet{wen2023transformers} & IJCAI & -- & -- & -- & -- & $\circ$\\
\citet{wang2024deep} & arXiv & $\circ$ & $\circ$ & -- & -- & \checkmark\\
\citet{kim2025comprehensive} & AI Review & $\circ$ & $\circ$ & -- & -- & --\\
\citet{toner2024linear} & ICML & $\circ$ & -- & linear only & -- & \checkmark\\
\textbf{This survey} & -- & \checkmark & \checkmark & \checkmark & \checkmark & \checkmark\\
\bottomrule
\end{tabular}
\end{table}

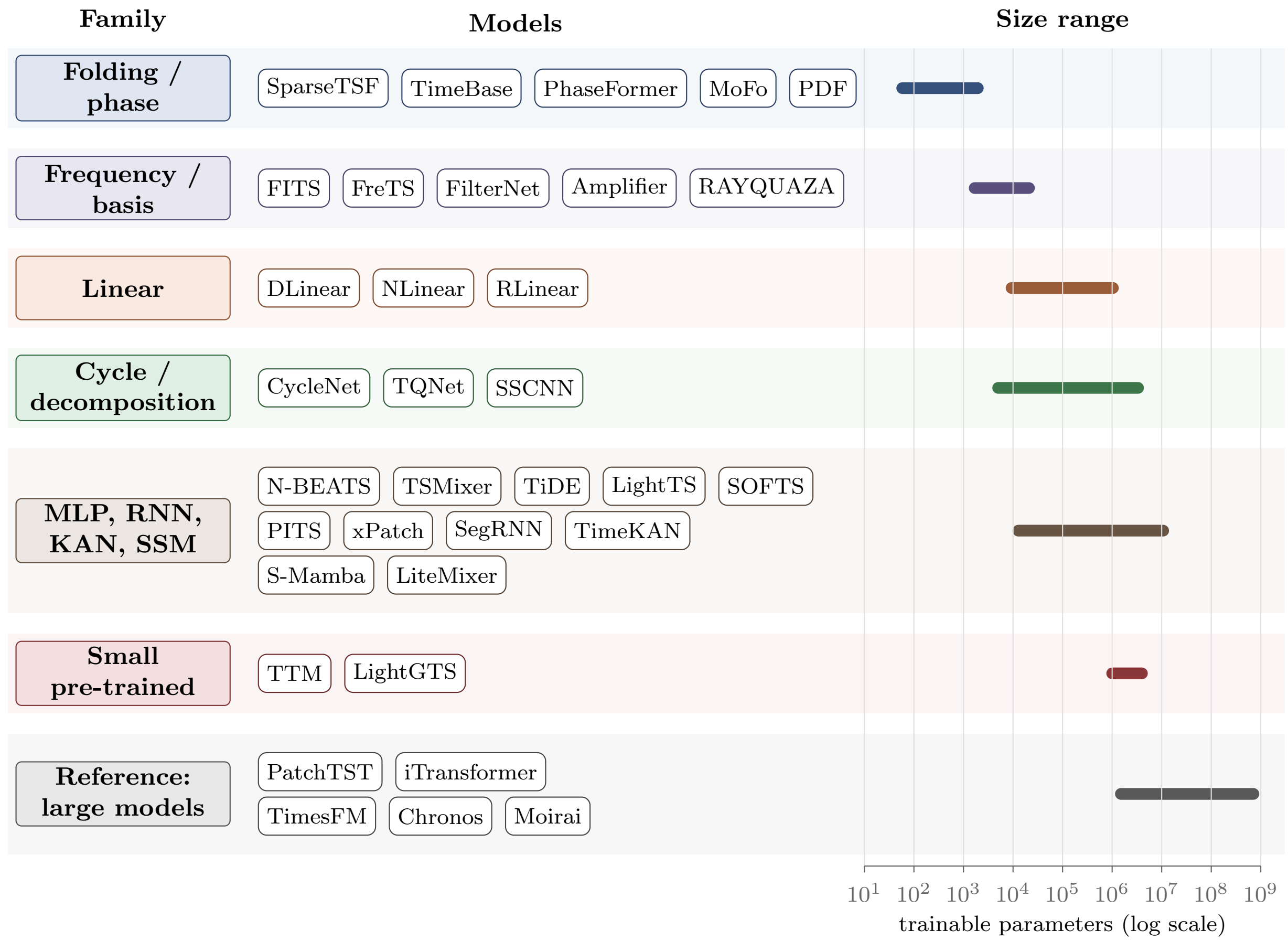
\begin{figure}[t]
\centering
\adjustbox{max width=\linewidth}{\begin{tikzpicture}[x=1cm,y=1cm,
fam/.style={draw=#1!65!black,fill=#1!18,rounded corners=2pt,minimum width=27mm,text width=25mm,minimum height=8mm,align=center,font=\footnotesize\bfseries,inner sep=2pt},
chip/.style={draw=#1!55!black,fill=white,rounded corners=3pt,font=\scriptsize,inner xsep=3pt,inner ysep=1.6pt,minimum height=4.8mm}]
\fill[cIn!6] (0,0.000) rectangle (16.100,-1.000);
\node[fam=cIn] at (1.45,-0.500) {Folding /\\phase};
\node[chip=cIn,anchor=west] (c1) at (3.150,-0.500) {SparseTSF};
\node[chip=cIn,anchor=west] (c2) at ([xshift=1.6mm]c1.east) {TimeBase};
\node[chip=cIn,anchor=west] (c3) at ([xshift=1.6mm]c2.east) {PhaseFormer};
\node[chip=cIn,anchor=west] (c4) at ([xshift=1.6mm]c3.east) {MoFo};
\node[chip=cIn,anchor=west] (c5) at ([xshift=1.6mm]c4.east) {PDF};
\draw[line width=4.2pt,cIn!70!black,line cap=round] (11.277,-0.500) -- (12.231,-0.500);
\fill[cFreq!6] (0,-1.260) rectangle (16.100,-2.260);
\node[fam=cFreq] at (1.45,-1.760) {Frequency /\\basis};
\node[chip=cFreq,anchor=west] (c6) at (3.150,-1.760) {FITS};
\node[chip=cFreq,anchor=west] (c7) at ([xshift=1.6mm]c6.east) {FreTS};
\node[chip=cFreq,anchor=west] (c8) at ([xshift=1.6mm]c7.east) {FilterNet};
\node[chip=cFreq,anchor=west] (c9) at ([xshift=1.6mm]c8.east) {Amplifier};
\node[chip=cFreq,anchor=west] (c10) at ([xshift=1.6mm]c9.east) {RAYQUAZA};
\draw[line width=4.2pt,cFreq!70!black,line cap=round] (12.191,-1.760) -- (12.877,-1.760);
\fill[cLin!6] (0,-2.520) rectangle (16.100,-3.520);
\node[fam=cLin] at (1.45,-3.020) {Linear};
\node[chip=cLin,anchor=west] (c11) at (3.150,-3.020) {DLinear};
\node[chip=cLin,anchor=west] (c12) at ([xshift=1.6mm]c11.east) {NLinear};
\node[chip=cLin,anchor=west] (c13) at ([xshift=1.6mm]c12.east) {RLinear};
\draw[line width=4.2pt,cLin!70!black,line cap=round] (12.656,-3.020) -- (13.935,-3.020);
\fill[cNorm!6] (0,-3.780) rectangle (16.100,-4.780);
\node[fam=cNorm] at (1.45,-4.280) {Cycle /\\decomposition};
\node[chip=cNorm,anchor=west] (c14) at (3.150,-4.280) {CycleNet};
\node[chip=cNorm,anchor=west] (c15) at ([xshift=1.6mm]c14.east) {TQNet};
\node[chip=cNorm,anchor=west] (c16) at ([xshift=1.6mm]c15.east) {SSCNN};
\draw[line width=4.2pt,cNorm!70!black,line cap=round] (12.487,-4.280) -- (14.252,-4.280);
\fill[cPar!6] (0,-5.040) rectangle (16.100,-7.120);
\node[fam=cPar] at (1.45,-6.080) {MLP, RNN,\\KAN, SSM};
\node[chip=cPar,anchor=west] (c17) at (3.150,-5.520) {N-BEATS};
\node[chip=cPar,anchor=west] (c18) at ([xshift=1.6mm]c17.east) {TSMixer};
\node[chip=cPar,anchor=west] (c19) at ([xshift=1.6mm]c18.east) {TiDE};
\node[chip=cPar,anchor=west] (c20) at ([xshift=1.6mm]c19.east) {LightTS};
\node[chip=cPar,anchor=west] (c21) at ([xshift=1.6mm]c20.east) {SOFTS};
\node[chip=cPar,anchor=west] (c22) at (3.150,-6.080) {PITS};
\node[chip=cPar,anchor=west] (c23) at ([xshift=1.6mm]c22.east) {xPatch};
\node[chip=cPar,anchor=west] (c24) at ([xshift=1.6mm]c23.east) {SegRNN};
\node[chip=cPar,anchor=west] (c25) at ([xshift=1.6mm]c24.east) {TimeKAN};
\node[chip=cPar,anchor=west] (c26) at (3.150,-6.640) {S-Mamba};
\node[chip=cPar,anchor=west] (c27) at ([xshift=1.6mm]c26.east) {LiteMixer};
\draw[line width=4.2pt,cPar!70!black,line cap=round] (12.742,-6.080) -- (14.569,-6.080);
\fill[cAtt!6] (0,-7.380) rectangle (16.100,-8.380);
\node[fam=cAtt] at (1.45,-7.880) {Small\\pre-trained};
\node[chip=cAtt,anchor=west] (c28) at (3.150,-7.880) {TTM};
\node[chip=cAtt,anchor=west] (c29) at ([xshift=1.6mm]c28.east) {LightGTS};
\draw[line width=4.2pt,cAtt!70!black,line cap=round] (13.925,-7.880) -- (14.301,-7.880);
\fill[gray!6] (0,-8.640) rectangle (16.100,-10.160);
\node[fam=gray] at (1.45,-9.400) {Reference:\\large models};
\node[chip=gray,anchor=west] (c30) at (3.150,-9.120) {PatchTST};
\node[chip=gray,anchor=west] (c31) at ([xshift=1.6mm]c30.east) {iTransformer};
\node[chip=gray,anchor=west] (c32) at (3.150,-9.680) {TimesFM};
\node[chip=gray,anchor=west] (c33) at ([xshift=1.6mm]c32.east) {Chronos};
\node[chip=gray,anchor=west] (c34) at ([xshift=1.6mm]c33.east) {Moirai};
\draw[line width=4.2pt,gray!70!black,line cap=round] (14.035,-9.400) -- (15.707,-9.400);
\draw[black!12] (10.800,0) -- (10.800,-10.310);
\draw[black!12] (11.425,0) -- (11.425,-10.310);
\draw[black!12] (12.050,0) -- (12.050,-10.310);
\draw[black!12] (12.675,0) -- (12.675,-10.310);
\draw[black!12] (13.300,0) -- (13.300,-10.310);
\draw[black!12] (13.925,0) -- (13.925,-10.310);
\draw[black!12] (14.550,0) -- (14.550,-10.310);
\draw[black!12] (15.175,0) -- (15.175,-10.310);
\draw[black!12] (15.800,0) -- (15.800,-10.310);
\draw[black!60] (10.800,-10.310) -- (15.800,-10.310);
\draw[black!60] (10.800,-10.310) -- ++(0,-0.08) node[below,font=\scriptsize] {$10^{1}$};
\draw[black!60] (11.425,-10.310) -- ++(0,-0.08) node[below,font=\scriptsize] {$10^{2}$};
\draw[black!60] (12.050,-10.310) -- ++(0,-0.08) node[below,font=\scriptsize] {$10^{3}$};
\draw[black!60] (12.675,-10.310) -- ++(0,-0.08) node[below,font=\scriptsize] {$10^{4}$};
\draw[black!60] (13.300,-10.310) -- ++(0,-0.08) node[below,font=\scriptsize] {$10^{5}$};
\draw[black!60] (13.925,-10.310) -- ++(0,-0.08) node[below,font=\scriptsize] {$10^{6}$};
\draw[black!60] (14.550,-10.310) -- ++(0,-0.08) node[below,font=\scriptsize] {$10^{7}$};
\draw[black!60] (15.175,-10.310) -- ++(0,-0.08) node[below,font=\scriptsize] {$10^{8}$};
\draw[black!60] (15.800,-10.310) -- ++(0,-0.08) node[below,font=\scriptsize] {$10^{9}$};
\node[font=\scriptsize,anchor=north] at (13.300,-10.780) {trainable parameters (log scale)};
\node[font=\footnotesize\bfseries,anchor=south] at (1.450,0.08) {Family};
\node[font=\footnotesize\bfseries,anchor=south] at (6.400,0.08) {Models};
\node[font=\footnotesize\bfseries,anchor=south] at (13.300,0.08) {Size range};
\end{tikzpicture}}
\caption{Taxonomy used in this survey. Rows: families; chips: the models discussed; bars: range of trainable parameters per shared predictor, from official code at $L=720$ in the ETTh1 configuration or as reported (SSCNN, pre-trained models; large models: \citet{das2024timesfm,ansari2024chronos,woo2024moirai}). FreTS (3.2M--24M, from code) lies outside the range shown for its family.}
\label{fig:taxonomy}
\Description{Taxonomy diagram with one row per model family (linear, frequency, period folding and phase, cycle and residual, light MLP and mixer, other designs, pre-trained). Each row lists its models as labelled chips and a horizontal bar on a log axis from ten to one billion showing the range of parameter counts of the family.}
\end{figure}

\paragraph{Contributions.}
\begin{enumerate}
\item \textbf{A structured-linear view} (Section~\ref{sec:theory}). We show that SparseTSF, TimeBase, FITS and DLinear are linear maps (affine after normalization) with Kronecker, band-limited, or unconstrained structure, extend the analysis of \citet{toner2024linear} to these constrained classes, give their parameter counts in closed form, and name the structured linear component of the non-linear designs (Table~\ref{tab:beyond}).
\item \textbf{A taxonomy of about 35 models} in six families (Figure~\ref{fig:taxonomy}, Sections~\ref{sec:families}--\ref{sec:pretrained}), with a summary table and an architecture diagram for each core model, and a discussion of the application domains where they matter (Section~\ref{sec:apps}).
\item \textbf{A code-level audit} (Section~\ref{sec:audit}). We count parameters from sixteen official repositories plus the Time-Series-Library, compare them with published numbers, and check evaluation code.
\item \textbf{Controlled experiments} (Section~\ref{sec:closedform}) that fit each structural constraint by least squares under one protocol, against a zero-parameter seasonal naive baseline.
\item \textbf{Practical guidance and open problems}: deployment considerations, a model-selection guide and a reporting protocol (Section~\ref{sec:practice}), and open problems (Section~\ref{sec:open}).
\end{enumerate}

\paragraph{What each part covers.} The parts of this survey cover different
sets of models (Table~\ref{tab:coverage}). The taxonomy is the widest. The
structured-linear view gives exact reductions only for the linear models
and names the linear component of the others. The audit covers sixteen
third-party repositories, and the controlled experiments cover the three
linear constraint classes on eight datasets (the four ETT
datasets, Electricity, Traffic, Weather and Exchange).

\begin{table}[t]
\centering
\caption{Coverage of each part of the survey.}
\label{tab:coverage}
\small
\begin{tabular}{@{}>{\raggedright\arraybackslash}p{2.3cm}>{\raggedright\arraybackslash}p{6.4cm}>{\raggedright\arraybackslash}p{3.6cm}@{}}
\toprule
Part & Models covered & Evidence\\
\midrule
Taxonomy (Sec.~\ref{sec:linear}--\ref{sec:pretrained}) & about 35 models in six families, the sixth a residual group, and lightweight pre-trained models & published papers and code\\
Structured-linear view (Sec.~\ref{sec:theory}) & exact reduction: Linear, DLinear, FITS, SparseTSF (linear), TimeBase, CycleNet-Linear. Linear component only: TQNet, PhaseFormer and the designs of Sec.~\ref{sec:mlp} and~\ref{sec:other} (Table~\ref{tab:beyond}) & derivations, numerical checks (App.~\ref{app:counting})\\
Code audit (Sec.~\ref{sec:audit}) & 16 third-party repositories and the Time-Series-Library reference models; selection in Sec.~\ref{sec:audit} & parameter counts and test loaders, from code\\
Controlled experiments (Sec.~\ref{sec:closedform}) & full, low-pass input and phase-shared linear maps, seasonal naive; ETTh1, ETTh2, ETTm1, ETTm2, Electricity, Traffic, Weather, Exchange & closed-form least squares, no network trained\\
Accuracy comparison (Table~\ref{tab:acc}) & models reported by \citet[Table~1]{niu2026phaseformer} and the released CycleNet logs & published results\\
\bottomrule
\end{tabular}
\end{table}

\paragraph{Scope and method.} We cover point forecasting of multivariate
series with look-back $L$ and horizon $H$. We call a model
\emph{lightweight} if its parameter count per shared predictor is at most
that of one linear map, $O(LH)$. We read the papers in full where
available (PMLR PDFs, arXiv HTML), cloned the official repositories, and
instantiated each model's constructor to count parameters
(Appendix~\ref{app:counting}). Every number in this paper is
labelled with its source: computed from the official code, or the paper it comes from.

\paragraph{Paper selection.} We started from the lightweight models most
cited as efficiency baselines (DLinear, FITS, SparseTSF) and followed the
baselines and related work of every included paper, forward and backward,
in the proceedings of NeurIPS, ICML, ICLR, AAAI and KDD, TMLR, and arXiv
(2021--2026). We complemented this with targeted searches for
\emph{lightweight}, \emph{efficient}, \emph{parameter-efficient} and
\emph{periodic} long-term forecasting. A model
is included if it (i) targets long-term point forecasting on the standard
benchmarks, and (ii) either claims efficiency as a main contribution or is
a standard efficiency baseline. Models whose official code is public are
also audited (Section~\ref{sec:audit}). The selection is curated rather
than exhaustive; Table~\ref{tab:master} lists every model we discuss.

\paragraph{Organization.} Section~\ref{sec:setting} describes the
forecasting setting, the benchmarks and the channel strategies.
Section~\ref{sec:theory} develops the structured-linear view.
Section~\ref{sec:families} reviews the six model families,
Section~\ref{sec:pretrained} the lightweight pre-trained models and
Section~\ref{sec:apps} the application domains. Section~\ref{sec:audit}
reports the code-level audit and Section~\ref{sec:closedform} the controlled
experiments. Section~\ref{sec:practice} turns the evidence into practical
guidance, Section~\ref{sec:open} lists open problems and
Section~\ref{sec:conclusion} concludes. The appendices give the counting
method, a table of all models and the official repositories.

\section{Background: How Deep Long-Term Forecasting Works}
\label{sec:setting}

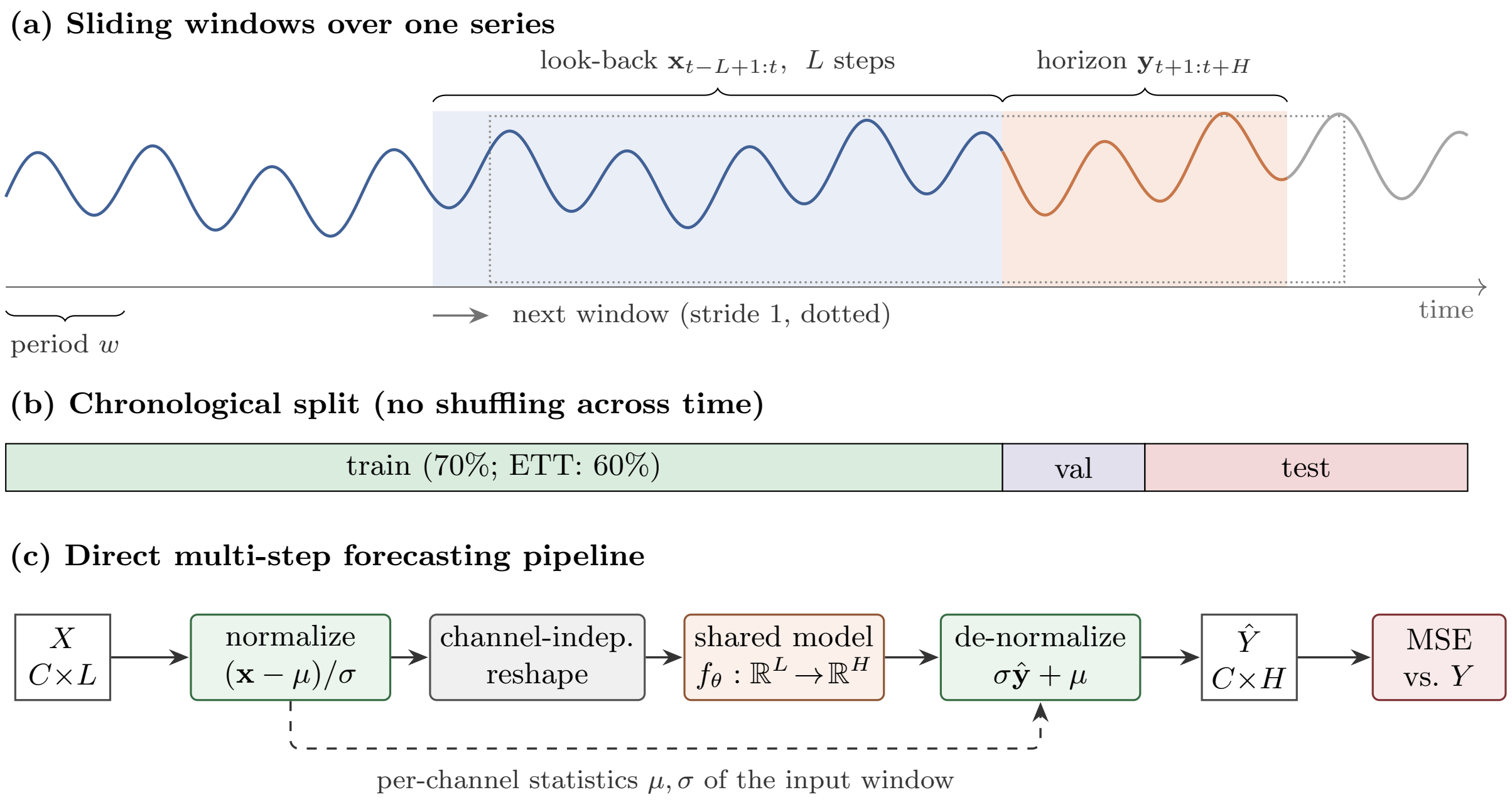
\begin{figure}[t]
\centering
\adjustbox{max width=\linewidth}{\begin{tikzpicture}[x=1cm,y=1cm]
\def\sig{0.95+0.38*sin(deg(2*pi*\x/1.25))+0.16*sin(deg(2*pi*\x/4.1))+0.03*\x}
\node[sub] at (-0.1,2.75) {(a) Sliding windows over one series};
\fill[cIn!12] (4.5,0) rectangle (10.5,1.85);
\fill[cLin!18] (10.5,0) rectangle (13.5,1.85);
\draw[black!45,densely dotted,line width=0.6pt] (5.1,0.05) rectangle (14.1,1.8);
\draw[->,black!60] (0,0) -- (15.6,0) node[below left] {\scriptsize time};
\draw[cIn!85!black,line width=0.9pt,domain=0:10.5,samples=300,smooth] plot (\x,{\sig});
\draw[cLin!90!black,line width=0.9pt,domain=10.5:13.5,samples=90,smooth] plot (\x,{\sig});
\draw[black!35,line width=0.9pt,domain=13.5:15.4,samples=60,smooth] plot (\x,{\sig});
\draw[decorate,decoration={brace,amplitude=4pt}] (4.5,1.95) -- (10.5,1.95) node[midway,above=4pt,lbl] {look-back $\mathbf x_{t-L+1:t}$, \;$L$ steps};
\draw[decorate,decoration={brace,amplitude=4pt}] (10.5,1.95) -- (13.5,1.95) node[midway,above=4pt,lbl] {horizon $\mathbf y_{t+1:t+H}$};
\draw[arr,black!55] (4.5,-0.3) -- (5.1,-0.3); \node[lbl,anchor=west] at (5.2,-0.3) {next window (stride 1, dotted)};
\draw[decorate,decoration={brace,amplitude=3pt,mirror}] (0,-0.25) -- (1.25,-0.25) node[midway,below=3pt,lbl] {period $w$};
\node[sub] at (-0.1,-1.25) {(b) Chronological split (no shuffling across time)};
\draw[fill=cNorm!22] (0,-2.15) rectangle (10.5,-1.65) node[midway] {train (70\%; ETT: 60\%)};
\draw[fill=cFreq!22] (10.5,-2.15) rectangle (12.0,-1.65) node[midway] {val};
\draw[fill=cAtt!22] (12.0,-2.15) rectangle (15.4,-1.65) node[midway] {test};
\node[sub] at (-0.1,-2.85) {(c) Direct multi-step forecasting pipeline};
\node[dat] (x) at (0.6,-3.9) {$X$\\$C{\times}L$};
\node[blk=cNorm,minimum width=21mm] (n) at (3.0,-3.9) {normalize\\$(\mathbf x-\mu)/\sigma$};
\node[blk=gray,minimum width=19mm] (ci) at (5.6,-3.9) {channel-indep.\\reshape};
\node[blk=cLin,minimum width=21mm] (f) at (8.2,-3.9) {shared model\\$f_\theta:\R^{L}\!\to\!\R^{H}$};
\node[blk=cNorm,minimum width=21mm] (dn) at (10.9,-3.9) {de-normalize\\$\sigma\hat{\mathbf y}+\mu$};
\node[dat] (y) at (13.1,-3.9) {$\hat Y$\\$C{\times}H$};
\node[blk=cAtt,minimum width=14mm] (loss) at (15.1,-3.9) {MSE\\vs.\ $Y$};
\foreach \a/\b in {x/n,n/ci,ci/f,f/dn,dn/y,y/loss}\draw[arr] (\a)--(\b);
\draw[darr] (n.south) -- ++(0,-0.5) -| (dn.south);
\node[lbl,anchor=north] at (6.95,-4.95) {per-channel statistics $\mu,\sigma$ of the input window};
\end{tikzpicture}}
\caption{The standard LTSF pipeline. (a) Supervised pairs are cut from one
long series with a sliding window: $L$ past steps as input, the next $H$
steps as target. (b) The series is split chronologically, never shuffled
across time. (c) Each window is normalized per channel, mapped to all $H$
future steps at once by a model shared across channels, and
de-normalized. The loss is the MSE on the normalized scale.}
\label{fig:pipeline}
\Description{Three stacked panels. Top: a wavy series with two overlapping windows marked, each an input segment followed by a target segment, and a bracket showing one period. Middle: a time axis split into consecutive train, validation and test blocks. Bottom: a left-to-right chain of boxes from the input window through normalization, channel-independent reshaping and a backbone to the forecast, compared with the target.}
\end{figure}

This section explains the standard pipeline shared by every model in the
survey. Figure~\ref{fig:pipeline} summarizes it.

\subsection{From a series to supervised pairs}

A multivariate series is a matrix $S\in\R^{C\times T}$ with $C$ channels
(sensors, customers, roads) and $T$ time steps. For every time $t$, the
model sees the look-back window $X_t=S_{:,t-L+1:t}\in\R^{C\times L}$ and
predicts $Y_t=S_{:,t+1:t+H}\in\R^{C\times H}$. Windows are taken with
stride 1, so consecutive samples overlap in $L+H-1$ steps. Forecasting is
thus a regression problem $f:\R^{C\times L}\to\R^{C\times H}$ whose samples
are strongly dependent.

\paragraph{Direct multi-step output.} Almost all models in this survey
predict the $H$ future values in one forward pass (direct multi-step),
instead of feeding predictions back one step at a time. This avoids error
accumulation and makes the output layer a map to $\R^{H}$, which is why
the horizon $H$ appears in every parameter count below.

\paragraph{Splits and scaling.} The series is split in time: 12/4/4 months
for the ETT datasets (60/20/20\% of the first 20 months) and 70/10/20\% for
the others
\citep{lin2024sparsetsf,huang2025timebase}. Each channel is standardized
with the mean and standard deviation of the training part, and the
reported MSE and MAE are computed on this standardized scale. The model
with the best validation loss is kept (early stopping), and the test
metrics are averaged over $H\in\{96,192,336,720\}$.

\paragraph{Benchmarks.} Table~\ref{tab:data} lists the datasets used by
nearly all lightweight papers. Their dominant periods matter for the
folding models of Section~\ref{sec:folding}: hourly data have a daily
period $w=24$ (and a weekly one, $168$, on Traffic and Electricity); 15-min
and 10-min data have daily periods of 96 and 144 steps. Exchange has no
stable period and is a useful negative control.

\begin{table}[t]
\centering
\caption{Standard LTSF benchmarks. Statistics from
\citet[Tables~1, 8]{lin2024sparsetsf} and \citet[Table~1]{huang2025timebase}.}
\label{tab:data}
\small
\begin{tabular}{lrlrll}
\toprule
Dataset & Channels & Sampling & Length & Dominant period & Split\\
\midrule
ETTh1, ETTh2 & 7 & 1 hour & 17{,}420 & 24 (day) & 6:2:2\\
ETTm1, ETTm2 & 7 & 15 min & 69{,}680 & 96 (day) & 6:2:2\\
Weather & 21 & 10 min & 52{,}696 & 144 (day) & 7:1:2\\
Electricity & 321 & 1 hour & 26{,}304 & 24 (day), 168 (week) & 7:1:2\\
Traffic & 862 & 1 hour & 17{,}544 & 24 (day), 168 (week) & 7:1:2\\
Exchange & 8 & 1 day & 7{,}588 & none & 7:1:2\\
\bottomrule
\end{tabular}
\end{table}

\subsection{Channel strategy}

A \emph{channel-dependent} (CD) model reads all $C$ channels jointly. A
\emph{channel-independent} (CI) model treats each channel as a separate
univariate sample and applies one shared function $f_\theta:\R^{L}\to\R^{H}$
to all of them; in code, the batch of shape $(B,C,L)$ is reshaped to
$(BC,L)$. CI has two consequences that matter here: $|\theta|$ does not
depend on $C$, and every channel acts as extra training data for the same
parameters. \citet{han2024capacity} analyse this as a trade-off between the
capacity of CD models and the robustness of CI models under distribution
shift. Most models in this survey are CI; CycleNet adds $wC$
channel-specific parameters, and TQNet and SSCNN add explicit channel
interaction.

\subsection{Instance normalization}

Real series drift: the level and scale of a window differ from those of the
training set. Reversible instance normalization (RevIN)
\citep{kim2022revin} removes the statistics of each window before the model
and restores them after:
\begin{equation}
\tilde{\mathbf x}=\frac{\mathbf x-\mu(\mathbf x)\mathbf 1}{\sigma(\mathbf x)},
\qquad
\hat{\mathbf y}=\sigma(\mathbf x)\,f_\theta(\tilde{\mathbf x})+\mu(\mathbf x)\mathbf 1 .
\label{eq:revin}
\end{equation}
Some models remove only the mean (SparseTSF). If $f_\theta$ is linear and
only the mean is removed, the full predictor is affine in $\mathbf x$. With
$\sigma(\mathbf x)$, it is positively homogeneous but not linear. This is
the only non-linearity in most ``linear'' forecasters. SparseTSF reports
that normalization matters on ETTh1/ETTh2 and has little effect on
Electricity and Traffic \citep[Table~13]{lin2024sparsetsf}.

\subsection{Tokenization: points, patches, phases, frequencies}

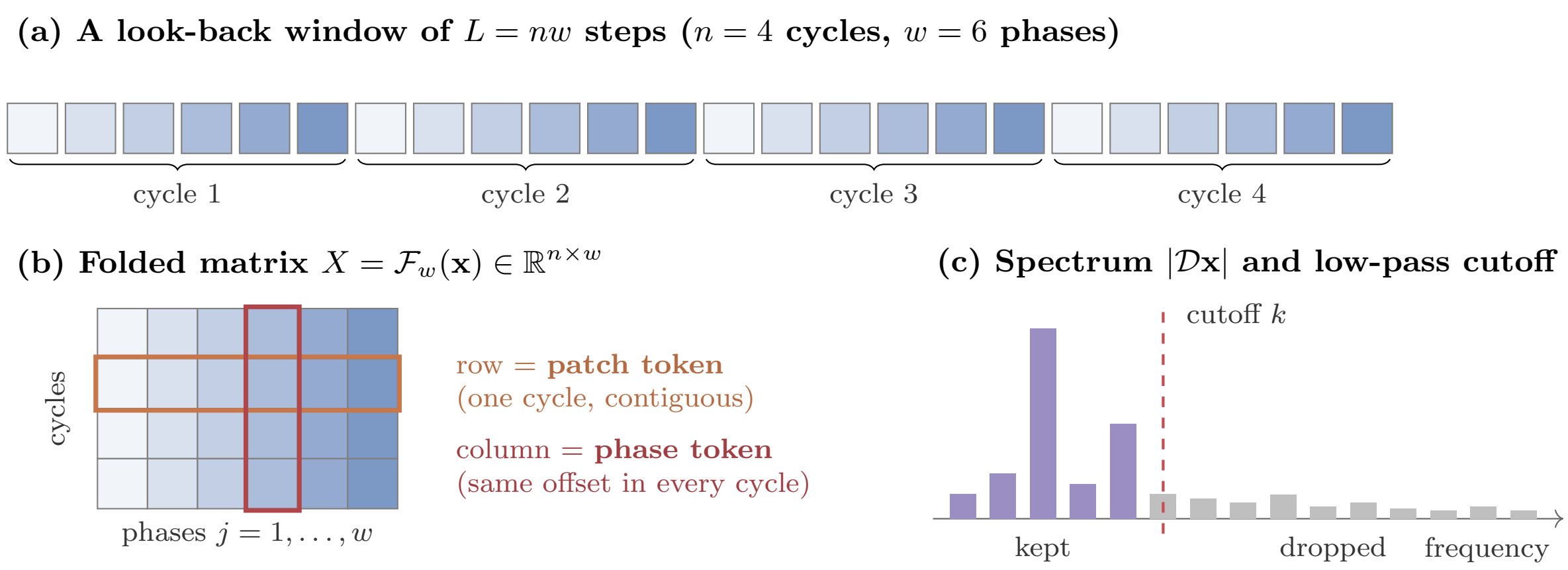
\begin{figure}[t]
\centering
\adjustbox{max width=\linewidth}{\begin{tikzpicture}[x=1cm,y=1cm]
\node[sub] at (-0.3,1.25) {(a) A look-back window of $L=nw$ steps ($n=4$ cycles, $w=6$ phases)};
\foreach \i in {0,...,23}{\pgfmathtruncatemacro{\ph}{mod(\i,6)}\pgfmathtruncatemacro{\sh}{8+\ph*13}
  \node[cell,minimum size=5mm,fill=cIn!\sh] (s\i) at (\i*0.58,0.3) {};}
\foreach \k in {0,1,2,3}{\pgfmathtruncatemacro{\a}{6*\k}\pgfmathtruncatemacro{\b}{6*\k+5}
  \draw[decorate,decoration={brace,amplitude=3pt,mirror}] ($(s\a.south west)+(0.03,-0.05)$) -- ($(s\b.south east)+(-0.03,-0.05)$) node[midway,below=3pt,lbl] {cycle \the\numexpr\k+1\relax};}
\node[sub] at (-0.3,-1.05) {(b) Folded matrix $X=\mathcal F_w(\mathbf x)\in\R^{n\times w}$};
\foreach \r in {0,...,3}{\foreach \c in {0,...,5}{\pgfmathtruncatemacro{\sh}{8+\c*13}
  \node[cell,minimum size=5mm,fill=cIn!\sh] (m\r\c) at (0.9+\c*0.5,-1.75-\r*0.5) {};}}
\node[draw=cLin!90!black,line width=1.3pt,fit=(m10)(m15),inner sep=0.3pt] {};
\node[draw=cAtt!90!black,line width=1.3pt,fit=(m03)(m33),inner sep=0.3pt] {};
\node[lbl,rotate=90] at (0.25,-2.5) {cycles};
\node[lbl] at (2.15,-3.75) {phases $j=1,\dots,w$};
\node[note,anchor=west,text=cLin!80!black] at (4.1,-2.25) {row = \textbf{patch token}\\(one cycle, contiguous)};
\node[note,anchor=west,text=cAtt!80!black] at (4.1,-3.1) {column = \textbf{phase token}\\(same offset in every cycle)};
\node[sub] at (8.9,-1.05) {(c) Spectrum $|\mathcal D\mathbf x|$ and low-pass cutoff};
\draw[->,black!60] (9.0,-3.6) -- (15.3,-3.6);
\node[lbl,anchor=north east] at (15.3,-3.65) {frequency};
\foreach \k/\h in {0/0.25,1/0.45,2/1.9,3/0.35,4/0.95,5/0.25,6/0.2,7/0.16,8/0.24,9/0.12,10/0.16,11/0.1,12/0.08,13/0.12,14/0.08}{
  \pgfmathsetmacro{\xx}{9.3+\k*0.4}
  \ifnum\k<5 \fill[cFreq!80] (\xx-0.13,-3.6) rectangle (\xx+0.13,-3.6+\h);
  \else \fill[black!25] (\xx-0.13,-3.6) rectangle (\xx+0.13,-3.6+\h); \fi}
\draw[dashed,cAtt,line width=0.8pt] (11.3,-3.75) -- (11.3,-1.45);
\node[lbl,anchor=west] at (11.4,-1.55) {cutoff $k$};
\node[lbl] at (10.1,-3.9) {kept};
\node[lbl] at (13.0,-3.9) {dropped};
\end{tikzpicture}}
\caption{Tokenizations. (a) A window of $n$ cycles of length $w$. (b)
Folding it into $X\in\R^{n\times w}$: a row is one cycle (patch token), a
column is one phase across cycles (phase token). (c) Frequency models keep
the $k$ lowest Fourier bins.}
\label{fig:tokens}
\Description{Three panels. (a) A look-back window made of several cycles of equal length. (b) The window folded into a matrix with one row per cycle and one column per phase, contrasting a patch token (a row segment) with a phase token (a column). (c) The magnitude spectrum of the window with a low-pass cutoff.}
\end{figure}

How the window is cut into pieces decides what a model can share, and
therefore its size. Figure~\ref{fig:tokens} shows the four choices.
\emph{Point} models (Linear, DLinear) see the raw $L$ values.
\emph{Patch} models (PatchTST, Crossformer) cut the window into contiguous
segments. \emph{Phase} models gather the values at the same offset of
every cycle. \emph{Frequency} models work on Fourier coefficients. Phases
are formalized by folding.

\begin{definition}[Period folding]
Let $w$ divide $L$ and $H$, with $n=L/w$ and $m=H/w$. Folding maps
$\mathbf x\in\R^{L}$ to $X=\mathcal F_w(\mathbf x)\in\R^{n\times w}$,
$X_{i,j}=x_{iw+j}$. Row $i$ is the $i$-th cycle; column $j$ is phase $j$
across cycles. With row-major vectorization, $\mathbf x=\vecr(X)$.
\end{definition}

\subsection{Training and evaluation}

Models are trained with the MSE loss and Adam, with early stopping on the
validation set. Test errors are
$\mathrm{MSE}=\frac{1}{CH}\sum_{c,h}(\hat y_{c,h}-y_{c,h})^2$ and
$\mathrm{MAE}=\frac{1}{CH}\sum_{c,h}|\hat y_{c,h}-y_{c,h}|$, averaged over all
test windows. Two details change the numbers more than one might expect:
the look-back $L$ (Section~\ref{sec:audit}, Table~\ref{tab:lookback}) and a
data-loader bug that drops the last test batch (Table~\ref{tab:droplast}).

\section{A Structured-Linear View of Lightweight Forecasting}
\label{sec:theory}

\paragraph{Relation to prior analysis.} \citet{toner2024linear} showed that
several popular linear forecasters (including DLinear, NLinear, RLinear and
FITS) are functionally equivalent to unconstrained linear regression on
suitably augmented features, admit closed-form solutions, and that these
closed-form solutions forecast better in most of their settings. We build on
that result in two directions. First, we characterize the \emph{constrained}
classes that the newer folding models use, which are strict subspaces of the
linear class rather than equivalent to it (Section~\ref{sec:closedform} shows
what the constraint costs or saves). Second, we tie each class to its
parameter count as implemented in the official code (Section~\ref{sec:audit}).

The reference point is the full linear map $\hat{\mathbf y}=W\mathbf x+\mathbf b$,
$W\in\R^{H\times L}$, with $LH+H$ parameters (519{,}120 at $L=H=720$).

\subsection{Phase sharing is a Kronecker constraint}

\begin{proposition}[Phase-shared maps]
\label{prop:kron}
Let $A\in\R^{m\times n}$ act on every phase column of $X$, $Y=AX$. Then
$\vecr(Y)=(A\otimes I_w)\,\mathbf x$.
\end{proposition}
\begin{proof}
Write $\mathbf x=\sum_i \mathbf e_i\otimes X_{i,:}$. Then
$(A\otimes I_w)\mathbf x=\sum_i (A\mathbf e_i)\otimes X_{i,:}
=\sum_k \mathbf e_k\otimes \sum_i A_{ki}X_{i,:}=\vecr(AX)$.
\end{proof}

The parameter count drops from $LH$ to $nm=LH/w^2$, a factor $w^2$ ($576$
for hourly data with $w=24$).

\begin{corollary}[SparseTSF]
\label{cor:sparsetsf}
SparseTSF \citep{lin2024sparsetsf} first applies a residual smoothing
convolution, $\mathbf x\mapsto(I+K)\mathbf x$, where $K$ is a Toeplitz
matrix built from a kernel of size $2\lfloor w/2\rfloor+1$, then a
phase-shared linear map. Its forecaster (before de-normalization) is
\begin{equation}
W_{\text{SparseTSF}}=(A\otimes I_w)(I+K),\qquad
|\theta|=nm+2\lfloor w/2\rfloor+1 .
\end{equation}
\end{corollary}
The count matches Theorem~3.1 of \citet{lin2024sparsetsf}. The official
code has no bias in either layer; the count from code is $925$ at
$L=H=720$, $w=24$.

\subsection{TimeBase in the same view}

TimeBase \citep{huang2025timebase} applies two stacked linear layers to
each phase vector, a basis extraction $\R^{n}\to\R^{R}$ and a segment
forecast $\R^{R}\to\R^{m}$. Both act on phase vectors, so TimeBase is a
phase-shared model in the sense of Proposition~\ref{prop:kron}, without the
smoothing convolution of SparseTSF. Its parameter count is TimeBase's
Theorem~3.1, $\tfrac{R}{P}L+\tfrac{R+1}{P}H+R$ with $P=w$: at $L=H=720$,
$w=24$, $R=6$ it gives $396$, the ``0.39K'' of the paper and the count from
code. Its orthogonality loss acts on the basis coefficients and regularizes
training.

\subsection{Frequency truncation gives a band-limited map}

FITS \citep{xu2024fits} applies an rFFT to the normalized input, keeps the
first $k$ bins, maps them with a complex linear layer to $k'=\lfloor k(L+H)/L\rfloor$
bins, zero-pads, and applies an inverse rFFT of length $L+H$. Every step is
linear, so the output (backcast and forecast together) is
$\hat{\mathbf z}=\mathcal D^{-1}P^\top\Lambda P\mathcal D\,\tilde{\mathbf x}$
with $P$ a truncation and $\Lambda\in\mathbb C^{k'\times k}$: the input
is seen only through its first $k$ Fourier bins. In the official code, $k=(\lfloor L/T\rfloor+1)\,h+10$
where $T$ is the base period and $h$ the number of harmonics. The layer has
$k k'+k'$ complex weights, that is $2(kk'+k')$ real parameters.

\begin{remark}[Counting convention]
\citet[Table~4]{lin2024sparsetsf} report 840 parameters for FITS at
$L=H=96$. The count from code is 1{,}680. The factor 2 is exactly the
difference between counting complex weights as one or two parameters.
Papers should state the convention.
\end{remark}

\subsection{Decomposition does not enlarge a linear class}

DLinear \citep{zeng2023dlinear} maps the trend $M\mathbf x$ (moving
average) and the remainder $(I-M)\mathbf x$ with two linear layers. Since $M$
is linear, $W_s(I-M)+W_tM$ is a single matrix in $\R^{H\times L}$. The
decomposition changes the parametrization, and hence optimization and
implicit regularization, but not the hypothesis class. It doubles the
parameter count (1{,}038{,}240 at $L=H=720$).

\subsection{Why folding loses little: what the literature reports}

Two published arguments explain why phase sharing costs little on periodic
data. SparseTSF shows that if $x=P+T$ with $P(t)=P(t+w)$, the periodic part
is constant along each phase column, so a phase-shared map only has to
forecast the trend \citep[Theorem~3.4]{lin2024sparsetsf}. PhaseFormer
proves, with Wedin's $\sin\Theta$ theorem \citep{wedin1972perturbation},
that when every cycle follows a common pattern, the phase (column)
subspace of the folded matrix is invariant to a change of that pattern, up
to noise, while the patch (row) subspace moves
\citep[Theorem~1]{niu2026phaseformer}. Empirically, PhaseFormer reports that
two principal components explain over 90\% of the variance of phase tokens
on Traffic, against more than eleven for patch tokens. Together, these
results explain why phase-wise predictors can be tiny and robust when the
period is stable, and why they degrade when it is not.

\subsection{Designs outside the linear class}
\label{sec:beyond}

The reduction above is exact only for Linear, DLinear, FITS, SparseTSF
(linear variant), TimeBase and CycleNet-Linear. TQNet, PhaseFormer and the
designs of Sections~\ref{sec:mlp} and~\ref{sec:other} are non-linear, so no
single matrix $W$ describes them. Each of them still contains a structured
linear map, and that map is where its weight sharing comes from.
Table~\ref{tab:beyond} names this component and the operation that breaks
linearity. The table explains where the parameters of each design go, but it
gives no closed-form forecaster. The controlled experiments of
Section~\ref{sec:closedform} therefore cover only the linear models.

\begin{table}[t]
\centering
\caption{Non-linear lightweight designs: the structured linear component each
contains and the operation that makes the forecaster non-linear.}
\label{tab:beyond}
\small
\begin{tabular}{@{}>{\raggedright\arraybackslash}p{2.4cm}>{\raggedright\arraybackslash}p{2.6cm}>{\raggedright\arraybackslash}p{4.3cm}>{\raggedright\arraybackslash}p{3.0cm}@{}}
\toprule
Design & Models & Structured linear part & Non-linear part\\
\midrule
Folding, non-linear predictor & SparseTSF-MLP, PhaseFormer & first layer shared by all phases, of the Kronecker form of Section~\ref{sec:theory} & ReLU (SparseTSF-MLP), routing attention across phases (PhaseFormer)\\
Cycle, channel attention & TQNet & look-back embedding $L\to d$ shared by channels, and learned cycle queries & attention across channels, MLP\\
MLP mixers & LightTS, TSMixer, TiDE, TimeMixer & maps over time shared by channels, and over channels shared by time steps & pointwise activations\\
Basis expansion & N-BEATS & read-out $\hat{\mathbf y}=V\boldsymbol\theta$ on a fixed or learned basis $V$ & MLP producing $\boldsymbol\theta$\\
Patches and segments & PatchTST, PITS, xPatch, SegRNN & one embedding shared by all patches, $I_n\otimes E$ for non-overlapping patches (linear in PatchTST and SegRNN) & attention (PatchTST), MLP (PITS), CNN (xPatch), GRU (SegRNN)\\
Convolution & ModernTCN & depthwise convolutions, i.e.\ Toeplitz maps along time & activations, feed-forward layers\\
Koopman & Koopa & linear operator advancing a learned latent state & learned encoder and decoder\\
State space & S-Mamba & linear state recurrence over variate tokens & input-dependent (selective) parameters\\
Kolmogorov--Arnold & TimeKAN & linear combination of fixed Chebyshev polynomials of each input & the polynomial features\\
Channel mixing & SOFTS & embedding $L\to d$ of each channel & MLP and pooling into a shared core\\
\bottomrule
\end{tabular}
\end{table}

\subsection{Summary}

\begin{figure}[t]
\centering
\includegraphics[width=\textwidth]{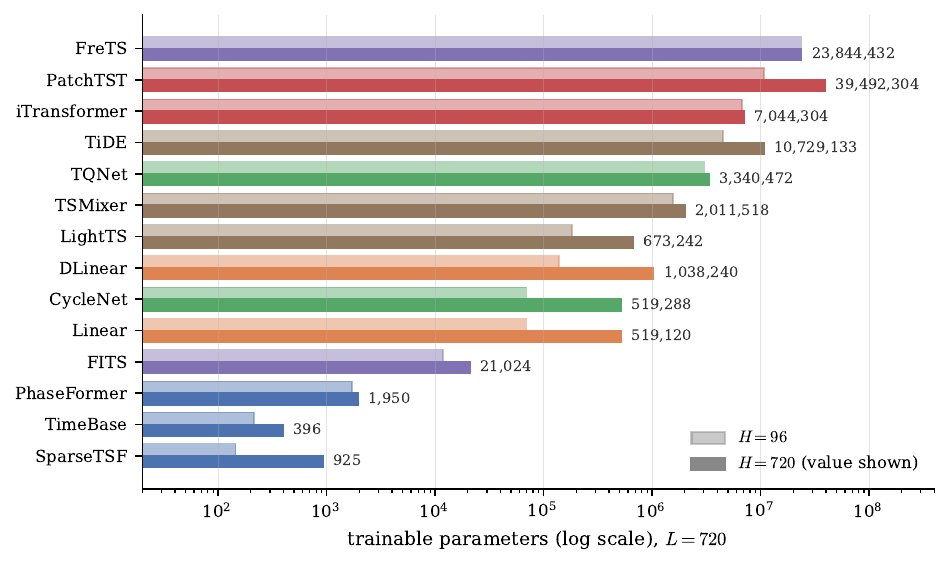}
\caption{Trainable parameters computed from the official constructors, ETTh1 configuration ($C=7$, $w=24$), look-back $L=720$. Light bar: $H=96$; dark bar: $H=720$ (value printed). Colors follow the families of Figure~\ref{fig:taxonomy}. Reference models use Time-Series-Library defaults. Log scale.}
\label{fig:sweep}
\Description{Line plots of trainable parameters on a log scale against look-back length, one line per lightweight model, in the ETTh1 configuration.}
\end{figure}

Table~\ref{tab:structure} lists each linear family as a constraint on
$W$, with the parameter formula of TQNet and PhaseFormer for comparison.
Figure~\ref{fig:sweep} compares the counts across models.

\begin{table}[t]
\centering
\caption{Lightweight models as constraints on the linear map. $n=L/w$,
$m=H/w$; $k,k'$ retained frequencies; $R$ number of basis components; $C$ channels; $d$
hidden width. Counts are exact for the official code (biases included).}
\label{tab:structure}
\small

\begin{tabular}{@{}l>{\raggedright\arraybackslash}p{3.6cm}>{\raggedright\arraybackslash}p{1.9cm}l@{}}
\toprule
Model & Operator (before de-normalization) & Constraint on $W$ & Parameters \\
\midrule
Linear & $W\mathbf x+\mathbf b$ & none & $LH+H$\\
DLinear \citep{zeng2023dlinear} & $W_s(I-M)\mathbf x+W_tM\mathbf x$ & none & $2(LH+H)$\\
FITS \citep{xu2024fits} & $\mathcal D^{-1}P^\top\Lambda P\mathcal D\mathbf x$ & band-limited ($k$ bins) & $2(kk'+k')$ real\\
SparseTSF \citep{lin2024sparsetsf} & $(A\otimes I_w)(I+K)\mathbf x$ & Kronecker & $nm+2\lfloor w/2\rfloor+1$\\
TimeBase \citep{huang2025timebase} & basis $n{\to}R$, then $R{\to}m$, per phase & phase-shared & $R(n+m)+R+m$\\
CycleNet-Linear \citep{lin2024cyclenet} & $W(\mathbf x-\mathbf q_{\text{in}})+\mathbf b+\mathbf q_{\text{out}}$ & none (affine shift) & $LH+H+wC$\\
TQNet \citep{lin2025tqnet} & attention over channels + MLP & non-linear & \makecell[l]{$4L^2{+}4L+wC+Ld{+}d$\\ ${}+2(d^2{+}d)+dH{+}H$}\\
PhaseFormer \citep{niu2026phaseformer} & phase embedding, routing, shared predictor & non-linear & $O(nd+wd+d^2+dm)$\\
\bottomrule
\end{tabular}
\end{table}

\section{Lightweight Model Families}
\label{sec:families}

This section reviews the models family by family, following
Figure~\ref{fig:taxonomy}. Each subsection opens with a summary table and
closes with a takeaway; Appendix~\ref{app:master} gives all models in one
table, and Appendix~\ref{app:code} lists the official code.

\subsection{Linear Models}
\label{sec:linear}

\begin{table}[t]
\centering
\caption{Linear models. Parameters as in Table~\ref{tab:master} ($L=720$, $H=96$, ETTh1 configuration unless marked).}
\label{tab:fam-linear}
\small
\begin{tabular}{llll}
\toprule
Model & Venue & Key idea & Params\\
\midrule
DLinear \citep{zeng2023dlinear} & AAAI'23 & moving-average split, two linear maps & 138.4K\\
NLinear \citep{zeng2023dlinear} & AAAI'23 & last-value subtraction, one linear map & 69.2K\\
RLinear \citep{li2023rlinear} & arXiv'23 & RevIN and one linear map & 69.2K\\
\bottomrule
\end{tabular}
\end{table}

\begin{figure}[t]
\centering
\adjustbox{max width=\linewidth}{\begin{tikzpicture}[x=1cm,y=1cm]
\node[dat] (x) at (0,0) {$\mathbf x$\\$\R^{L}$};
\node[blk=cNorm,minimum width=24mm] (ma) at (2.3,0) {moving-average\\decomposition};
\node[blk=cLin,minimum width=22mm] (lt) at (7.3,1.0) {Linear $W_t$\\$L\to H$};
\node[blk=cLin,minimum width=22mm] (ls) at (7.3,-1.0) {Linear $W_s$\\$L\to H$};
\node[op] (p) at (9.6,0) {$+$};
\node[dat] (y) at (10.9,0) {$\hat{\mathbf y}$\\$\R^{H}$};
\draw[arr] (x)--(ma);
\draw[line width=0.65pt,black!80] (ma.east) -- (3.75,0);
\draw[arr] (3.75,0) |- (lt.west);
\draw[arr] (3.75,0) |- (ls.west);
\fill[black!80] (3.75,0) circle (1.2pt);
\node[lbl,anchor=south] at (5.0,1.02) {trend $M\mathbf x$};
\node[lbl,anchor=north] at (4.75,-1.02) {remainder $(I{-}M)\mathbf x$};
\draw[arr] (lt.east) -| (p.north); \draw[arr] (ls.east) -| (p.south); \draw[arr] (p)--(y);
\node[note,anchor=west] at (11.7,0.25) {$\hat{\mathbf y}=\big(W_tM+W_s(I-M)\big)\mathbf x$};
\node[note,anchor=west] at (11.7,-0.3) {= one matrix in $\R^{H\times L}$};
\end{tikzpicture}}
\caption{DLinear \citep{zeng2023dlinear}. A moving average splits trend and remainder; each is mapped by its own linear layer. The sum is again a single linear map (Section~\ref{sec:theory}).}
\label{fig:dlinear}
\Description{Left-to-right block diagram. The input of length L enters a moving-average decomposition box, which splits into an upper path labelled trend and a lower path labelled remainder. Each path has one linear box from L to H; the two outputs meet at a plus node that produces the forecast y.}
\end{figure}
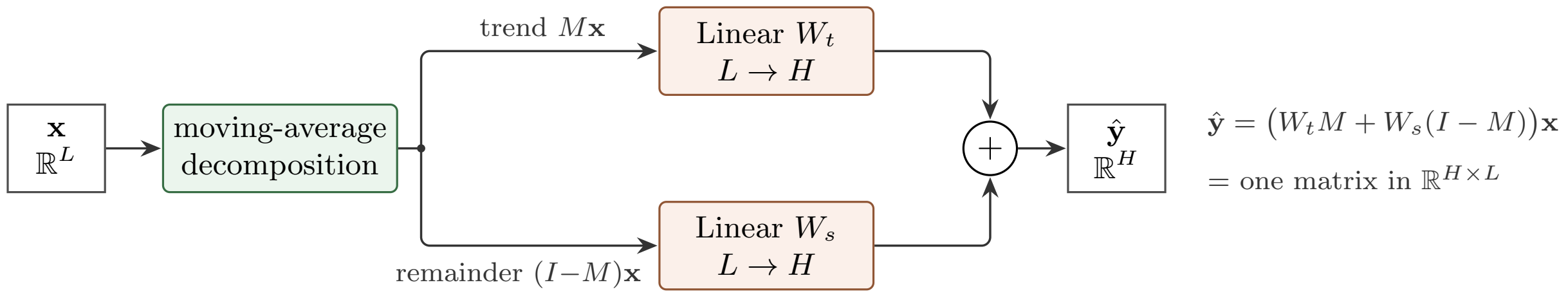

\paragraph{DLinear and NLinear.} \citet{zeng2023dlinear} (Figure~\ref{fig:dlinear}) questioned the
Transformer trend with two one-layer models. DLinear adds a moving-average
decomposition; NLinear subtracts the last value. By
Section~\ref{sec:theory}, both are a single linear map in hypothesis space.
Their parameter count, $O(LH)$, is the upper bound of our definition of
lightweight.

\paragraph{RLinear.} \citet{li2023rlinear} study a single linear layer with
RevIN and argue that linear mapping is effective on periodic data. Its
RevIN has an affine part, adding $2C$ parameters (14 on ETTh1).

\paragraph{What the linear family teaches.} SparseTSF visualizes the
learned $L\times H$ weights of a linear model and finds evenly spaced
stripes, that is, the model learns periodic structure by itself
\citep[Fig.~4]{lin2024sparsetsf}. The folding models below impose that
structure instead of learning it.

\subsection{Frequency and Basis Models}
\label{sec:frequency}

\begin{table}[t]
\centering
\caption{Frequency and basis models. Parameters as in Table~\ref{tab:master} ($L=720$, $H=96$, ETTh1 configuration unless marked).}
\label{tab:fam-freq}
\small
\begin{tabular}{llll}
\toprule
Model & Venue & Key idea & Params\\
\midrule
FITS \citep{xu2024fits} & ICLR'24 & low-pass, complex linear interpolation & 11.8K\\
FreTS \citep{yi2023frets} & NeurIPS'23 & MLPs in the frequency domain & 23.68M\\
FilterNet \citep{yi2024filternet} & NeurIPS'24 & learnable frequency filters & n.c.\\
Amplifier \citep{fei2025amplifier} & AAAI'25 & amplifies low-energy frequency bins & 55.1K$^\ast$\\
RAYQUAZA \citep{aityahia2026rayquaza} & OpenReview & per-window radial-basis atoms & n.c.\\
\bottomrule
\end{tabular}
\end{table}

\begin{figure}[t]
\centering
\adjustbox{max width=\linewidth}{\begin{tikzpicture}[x=1cm,y=1cm]
\node[dat] (x) at (0,0) {$\mathbf x$\\$\R^{L}$};
\node[blk=cNorm,minimum width=13mm] (n) at (1.75,0) {RIN};
\node[blk=cFreq,minimum width=13mm] (f) at (3.45,0) {rFFT};
\node[blk=cFreq,minimum width=19mm] (lp) at (5.55,0) {low-pass:\\keep $k$ bins};
\node[blk=cLin,minimum width=28mm] (cl) at (8.45,0) {complex Linear\\$\mathbb C^{k}\to\mathbb C^{k'}$};
\node[blk=cFreq,minimum width=17mm] (zp) at (11.3,0) {zero-pad to\\$(L{+}H)/2{+}1$};
\node[blk=cFreq,minimum width=17mm] (if) at (13.6,0) {irFFT};
\node[blk=cNorm,minimum width=17mm] (dn) at (13.6,-1.7) {iRIN, $\times\frac{L+H}{L}$};
\node[dat,minimum width=20mm] (y) at (11.0,-1.7) {$[\hat{\mathbf x};\hat{\mathbf y}]\in\R^{L+H}$};
\foreach \a/\b in {x/n,n/f,f/lp,lp/cl,cl/zp,zp/if,if/dn,dn/y}\draw[arr] (\a)--(\b);
\node[note,anchor=north west] at (0,-1.3) {$k=(\lfloor L/T\rfloor+1)\,h+10$ \;($T$: base period, $h$: harmonics)\\
$k'=\lfloor k(L+H)/L\rfloor$, \; $2(kk'+k')$ real parameters\\
all steps linear: one band-limited linear map};
\node[lbl,anchor=north] at (y.south) {backcast + forecast};
\end{tikzpicture}}
\caption{FITS \citep{xu2024fits}. Normalized input, rFFT, low-pass truncation to $k$ bins, one complex linear layer that interpolates the spectrum to the longer window $L{+}H$, zero-padding and inverse rFFT. The output contains the backcast and the forecast.}
\label{fig:fits}
\Description{Left-to-right chain of boxes: RIN, rFFT, a low-pass box keeping k bins, one complex linear box from k to k-prime bins, zero padding, irFFT and de-normalization. The output spans L plus H steps and is annotated as backcast plus forecast.}
\end{figure}
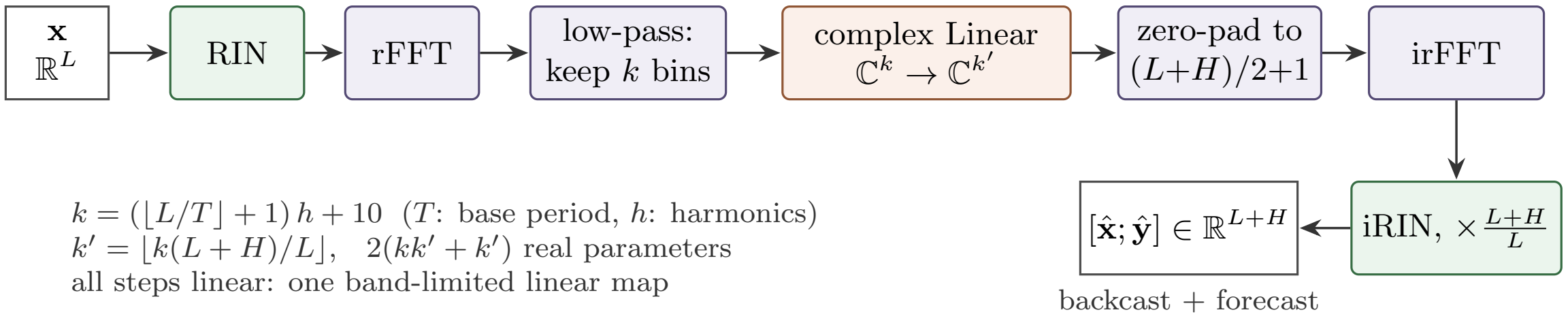

\paragraph{FITS.} FITS \citep{xu2024fits} (Figure~\ref{fig:fits}) brought linear forecasters to the
$10^4$ scale. It interpolates the low-frequency spectrum of the input to the
spectrum of the longer window $[\mathbf x;\mathbf y]$. Its cutoff controls
both accuracy and size. Counts from code with $h=2$ range from $1{,}680$
($L=H=96$) to $21{,}024$ ($L=H=720$).

\paragraph{FreTS and FilterNet.} FreTS \citep{yi2023frets} applies MLPs
in the frequency domain along the channel and time axes. Despite its
spectral design, it is not lightweight: its final projection maps an
$L\times128$ embedding to a hidden layer, which gives $3.2$M parameters at
$L=96$ and $23.8$M at $L=720$ (from code). FilterNet \citep{yi2024filternet}
learns frequency filters for robustness.

\paragraph{Amplifier.} \citet{fei2025amplifier} observe that forecasters
tend to ignore low-energy frequency components. An energy-amplification
block flips the spectrum so that low-energy components receive high energy,
a seasonal-trend forecaster models the two resulting spectral peaks, and an
energy-restoration block maps the output back. An optional
semi-channel-interaction block adds cross-channel structure. Its released
ETTh1 configuration without that block has 55.1K parameters (from code).

\begin{figure}[t]
\centering
\adjustbox{max width=\linewidth}{\begin{tikzpicture}[x=1cm,y=1cm]
\node[dat] (x) at (0,0) {$\mathbf x$\\$\R^{L}$};
\node[blk=cIn,minimum width=30mm] (g) at (2.7,0) {atom generator\\{\scriptsize (conditioned on $\mathbf x$)}};
\node[blk=cFreq,minimum width=36mm,minimum height=11mm] (rbf) at (8.1,0) {$\displaystyle\sum_{k=1}^{K} a_k\exp\!\Big(-\tfrac{(t-c_k)^2}{2s_k^2}\Big)$\\{\scriptsize input-conditioned RBF (iRBF)}};
\node[blk=cLin,minimum width=27mm] (fpl) at (12.2,0) {Forecast Projection\\Layer $\to\R^{H}$};
\node[dat] (y) at (14.6,0) {$\hat{\mathbf y}$\\$\R^{H}$};
\draw[arr] (x)--(g); \draw[arr] (g)--node[above,lbl]{$\{c_k,s_k,a_k\}$}(rbf); \draw[arr] (rbf)--(fpl); \draw[arr] (fpl)--(y);
\node[note,anchor=north west] at (0.2,-1.25) {fixed bases (Fourier, FITS) \quad vs.\quad global learned bases (TimeBase)\quad vs.\quad \textbf{per-window generated bases} (RAYQUAZA)};
\end{tikzpicture}}
\caption{RAYQUAZA \citep{aityahia2026rayquaza}. Gaussian atoms are
generated for each input window (iRBF) and a Forecast Projection Layer maps
the decomposition to the horizon.}
\label{fig:rayquaza}
\Description{The input window of length L feeds a generator box annotated conditioned on x, which outputs K triples of centre, width and amplitude. Each triple defines one Gaussian bump over the horizon; the bumps are summed to give the forecast y of length H.}
\end{figure}
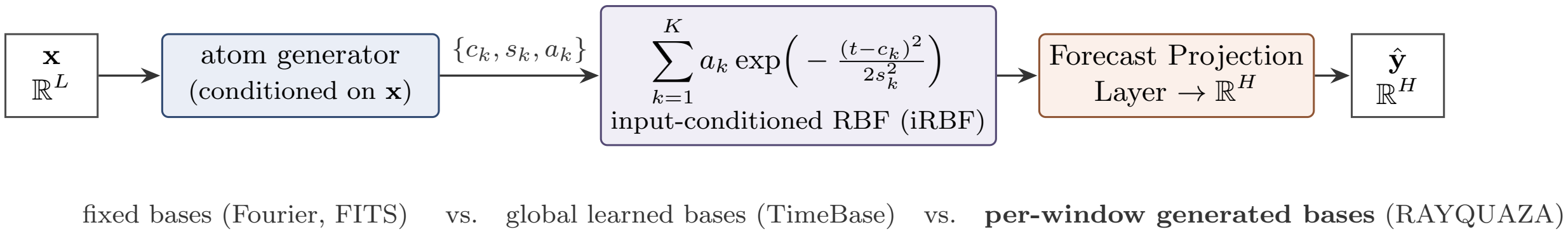

\paragraph{RAYQUAZA.} \citet{aityahia2026rayquaza} replace fixed
or globally learned bases with bases generated per input window
(Figure~\ref{fig:rayquaza}). An input-conditioned radial basis function
(iRBF) module produces, for each window, a set of Gaussian atoms with
their own centers $c_k$, widths $s_k$ and amplitudes $a_k$, and a Forecast
Projection Layer maps the resulting representation to the horizon. This
gives a three-way distinction for basis models: fixed bases (the Fourier
basis truncated by FITS), global learned bases shared by all windows
(TimeBase, Section~\ref{sec:folding}), and per-window generated bases
(RAYQUAZA). Localized atoms are designed to capture spikes and transients, which a
truncated Fourier basis spreads over many coefficients.

\paragraph{Takeaway.} A spectral view does not by itself make a model small.
FITS is small because it truncates; FreTS is not because it projects a
full-length embedding. Bases can be fixed, learned globally, or generated
per window; the last option trades a small generator for adaptivity.

\subsection{Period-Folding and Phase Models}
\label{sec:folding}

\begin{table}[t]
\centering
\caption{Period-folding and phase models. Parameters as in Table~\ref{tab:master} ($L=720$, $H=96$, ETTh1 configuration unless marked).}
\label{tab:fam-fold}
\small
\begin{tabular}{llll}
\toprule
Model & Venue & Key idea & Params\\
\midrule
SparseTSF \citep{lin2024sparsetsf} & ICML'24 & one linear map shared by all phases & 145\\
TimeBase \citep{huang2025timebase} & ICML'25 & segment basis per phase & 214\\
PhaseFormer \citep{niu2026phaseformer} & ICLR'26 & phase tokens, router attention & 1.7K$^\ddagger$\\
MoFo \citep{ma2025mofo} & NeurIPS'25 & period-structured 2D patches & n.c.\\
PDF \citep{dai2024pdf} & ICLR'24 & period-decoupled 2D variation & n.c.\\
\bottomrule
\end{tabular}
\end{table}

The models in this family fold the look-back window by its period $w$ and
share parameters across the $w$ phases (Section~\ref{sec:theory}). SparseTSF,
TimeBase and PhaseFormer are the smallest models in this survey, with about
one hundred to two thousand parameters.

\begin{figure}[t]
\centering
\adjustbox{max width=\linewidth}{\begin{tikzpicture}[x=1cm,y=1cm]
\node[dat] (x) at (0,0) {$\mathbf x$\\$\R^{L}$};
\node[blk=cNorm,minimum width=12mm] (n) at (1.55,0) {$-\,\bar x$};
\node[blk=cIn,minimum width=24mm] (cv) at (3.65,0) {$+$ Conv1D\\{\scriptsize kernel $2\lfloor w/2\rfloor{+}1$}};
\foreach \r in {0,...,3}{\foreach \c in {0,...,5}{\pgfmathtruncatemacro{\sh}{8+\c*13}
 \node[cell,minimum size=4.5mm,fill=cIn!\sh] (a\r\c) at (6.0+\c*0.45,0.68-\r*0.45) {};}}
\node[lbl,anchor=south] at ($(a02)!0.5!(a03)+(0,0.25)$) {fold: $n$ cycles $\times$ $w$ phases};
\draw[arr] (cv.east) -- (cv.east-|a10.west);
\node[blk=cLin,minimum width=34mm,minimum height=8mm] (lin) at ($(a32)!0.5!(a33)+(0,-1.35)$) {one Linear $A:\R^{n}\to\R^{m}$\\applied to every column};
\foreach \c in {0,...,5}{\draw[arr,cLin!80!black] (a3\c.south) -- (a3\c.south|-lin.north);}
\foreach \r in {0,...,2}{\foreach \c in {0,...,5}{\pgfmathtruncatemacro{\sh}{8+\c*13}
 \node[cell,minimum size=4.5mm,fill=cLin!\sh] (b\r\c) at (9.9+\c*0.45,0.45-\r*0.45) {};}}
\node[lbl,anchor=south] at ($(b02)!0.5!(b03)+(0,0.25)$) {$m\times w$};
\draw[arr] (lin.east) -| ($(b22)!0.5!(b23)+(0,-0.25)$);
\node[blk=cNorm,minimum width=15mm] (dn) at (13.45,0) {unfold\\$+\,\bar x$};
\node[dat] (y) at (15.05,0) {$\hat{\mathbf y}$\\$\R^{H}$};
\draw[arr] (x)--(n); \draw[arr] (n)--(cv); \draw[arr] (b15.east|-dn.west) -- (dn); \draw[arr] (dn)--(y);
\node[note,anchor=north west] at (11.2,-0.95) {$W=(A\otimes I_w)(I+K)$\\$nm+2\lfloor w/2\rfloor+1$ parameters\\(925 at $L=H=720$, $w=24$)};
\end{tikzpicture}}
\caption{SparseTSF \citep{lin2024sparsetsf}. After mean removal and a residual smoothing convolution, the window is folded into $n$ cycles by $w$ phases; one linear layer $A:\R^n\to\R^m$ is applied to every phase column, and the result is unfolded.}
\label{fig:sparsetsf}
\Description{The input window has its mean subtracted and a smoothing branch added back, then is drawn as a grid of n rows of cycles by w columns of phases. One shared arrow maps every column from n to m values, giving an m by w grid that is unfolded and has the mean added back. An annotation states the closed-form parameter count, 925 at L = H = 720 and w = 24.}
\end{figure}
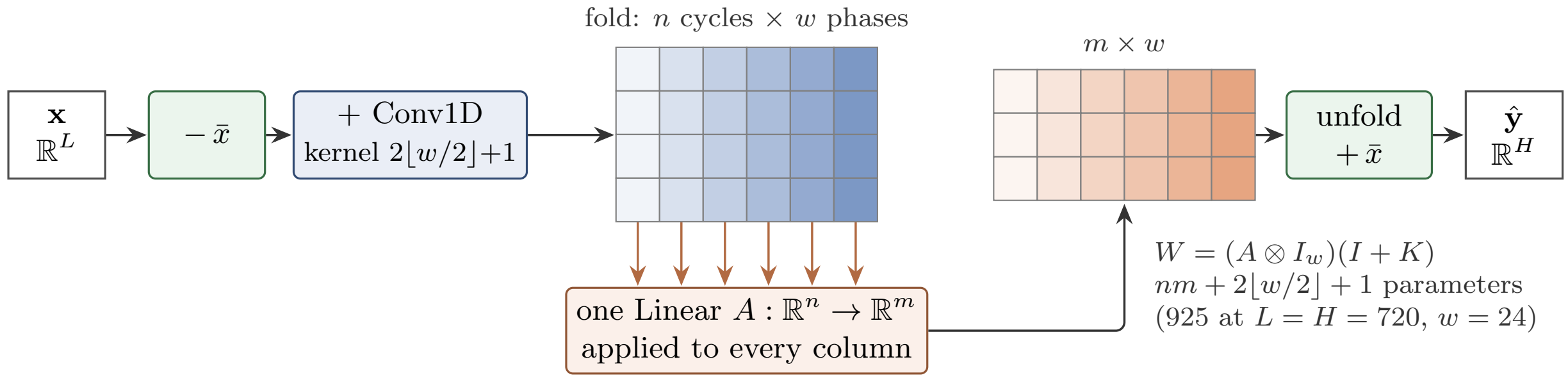

\paragraph{SparseTSF.} \citet{lin2024sparsetsf} (Figure~\ref{fig:sparsetsf}) downsample the smoothed
input with stride $w$ and apply one linear layer shared by all $w$ phases
(Corollary~\ref{cor:sparsetsf}). The paper shows that, if $x=P+T$ with
$P(t)=P(t+w)$, the periodic part is constant along each phase, so the model
only forecasts the trend \citep[Theorem~3.4]{lin2024sparsetsf}. It reports
three further findings. The period matters: on ETTh1 the average MSE is
$0.405$, $0.400$, $0.394$, $0.402$ for $w=6,12,24,48$. Some sparsity helps
even when it does not match the period: on Weather ($H=720$), $w=1$ (a full
linear layer) gives $0.322$ while $w=2$--$12$ give $0.318$--$0.319$. And
very long periods hurt, so the authors use $w=4$ on ETTm1/ETTm2/Weather
\citep[Tables~6, 9, 10]{lin2024sparsetsf}. These tables are from the
pre-fix protocol (Section~\ref{sec:audit}); the post-fix main results are
in its Appendix~D.1.

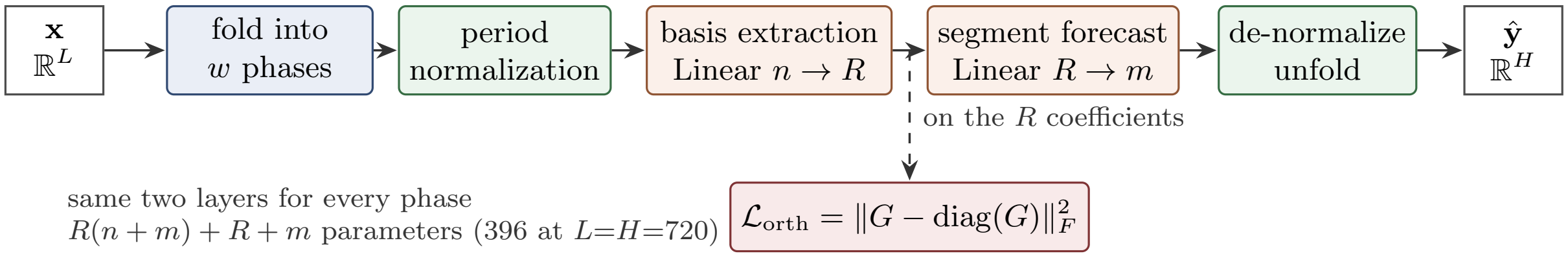
\begin{figure}[t]
\centering
\adjustbox{max width=\linewidth}{\begin{tikzpicture}[x=1cm,y=1cm]
\node[dat] (x) at (0,0) {$\mathbf x$\\$\R^{L}$};
\node[blk=cIn,minimum width=21mm] (fold) at (2.2,0) {fold into\\$w$ phases};
\node[blk=cNorm,minimum width=17mm] (pn) at (4.6,0) {period\\normalization};
\node[blk=cLin,minimum width=25mm] (v) at (7.3,0) {basis extraction\\Linear $n\to R$};
\node[blk=cLin,minimum width=25mm] (u) at (10.2,0) {segment forecast\\Linear $R\to m$};
\node[blk=cNorm,minimum width=18mm] (dn) at (12.9,0) {de-normalize\\unfold};
\node[dat] (y) at (14.9,0) {$\hat{\mathbf y}$\\$\R^{H}$};
\foreach \a/\b in {x/fold,fold/pn,pn/v,v/u,u/dn,dn/y}\draw[arr] (\a)--(\b);
\coordinate (mid) at ($(v.east)!0.5!(u.west)$);
\node[blk=cAtt,minimum width=36mm,minimum height=7mm] (orth) at (mid|-0,-1.7) {$\mathcal L_{\text{orth}}=\|G-\operatorname{diag}(G)\|_F^2$};
\draw[darr] (mid) -- node[right,lbl]{on the $R$ coefficients} (orth.north);
\node[note,anchor=west] at (0,-1.7) {same two layers for every phase\\$R(n+m)+R+m$ parameters (396 at $L{=}H{=}720$)};
\end{tikzpicture}}
\caption{TimeBase \citep{huang2025timebase}. Each phase vector is compressed to $R$ basis coefficients and expanded to $m$ future cycles by two linear layers without activation. An orthogonality penalty acts on the basis coefficients.}
\label{fig:timebase}
\Description{Left-to-right chain: fold into w phases, period normalization, a basis-extraction linear box from n to R, a segment-forecast linear box from R to m, then de-normalize and unfold to the forecast. An extra arrow marks the orthogonality loss applied to the R coefficients.}
\end{figure}

\paragraph{TimeBase.} \citet{huang2025timebase} (Figure~\ref{fig:timebase}) segment the series by
period, extract $R$ basis components with a linear layer, and forecast
whole segments. Both of its layers act on phase vectors, so it is a
phase-shared map (Section~\ref{sec:theory}). It reports $0.39$K parameters, $2.77$M MACs and $0.98$ms CPU inference
on Electricity ($L=H=720$), against $1.00$K, $12.71$M and $2.59$ms for
SparseTSF \citep[Table~3]{huang2025timebase}. It can also shrink
patch-based models: it reports 77.74--93.03\% fewer MACs for PatchTST.

\begin{figure}[t]
\centering
\adjustbox{max width=\linewidth}{\begin{tikzpicture}[x=1cm,y=1cm]
\node[dat] (x) at (0,0) {$\mathbf x$\\$\R^{L}$};
\node[blk=cIn,minimum width=20mm] (fold) at (1.95,0) {phase tokens\\$w\times\R^{n}$};
\node[blk=cLin,minimum width=16mm] (emb) at (4.2,0) {embed\\$n\to d$};
\node[blk=cAtt,minimum width=34mm,minimum height=7mm] (agg) at (8.0,1.35) {aggregate: phases $\to$ routers};
\node[blk=cAtt,minimum width=34mm,minimum height=7mm] (dis) at (8.0,0) {distribute: routers $\to$ phases};
\node[blk=gray,minimum width=34mm,minimum height=7mm] (mlp) at (8.0,-1.35) {residual + LN + MLP};
\node[grp,fit=(agg)(dis)(mlp),inner sep=6pt] (g) {};
\node[lbl,anchor=north] at (g.south) {cross-phase routing, $\times N$ layers};
\node[blk=cPar,minimum width=34mm,minimum height=6.5mm] (R) at (8.0,2.95) {$M$ routers $\in\R^{M\times d}$};
\node[blk=cLin,minimum width=17mm] (pred) at (12.1,0) {predictor\\$d\to m$};
\node[blk=cNorm,minimum width=16mm] (dn) at (14.15,0) {unfold\\iRevIN};
\node[dat] (y) at (15.9,0) {$\hat{\mathbf y}$\\$\R^{H}$};
\coordinate (br) at ($(g.west)+(-0.45,0)$);
\coordinate (ex) at ($(g.east|-mlp.east)+(0.35,0)$);
\draw[arr] (x)--(fold); \draw[arr] (fold)--(emb);
\draw[line width=0.65pt,black!80] (emb.east)--(br); \fill[black!80] (br) circle (1.4pt);
\draw[arr] (br) |- (agg.west);
\draw[arr] (br) -- (dis.west);
\draw[arr] (R)--(agg); \draw[arr] (agg)--(dis); \draw[arr] (dis)--(mlp);
\draw[arr] (mlp.east) -- (ex) |- (pred.west);
\draw[arr] (pred)--(dn); \draw[arr] (dn)--(y);
\node[lbl,anchor=south] at (pred.north) {shared by all phases};
\node[note,anchor=north west] at (0,-1.0) {RevIN at input and output omitted\\routing cost $O\big(N((w{+}M)d^2{+}Mwd)\big)$\\(full attention: $O(w^2d)$)\\$\approx1.7$K parameters with $d=8$, $N=1$};
\end{tikzpicture}}
\caption{PhaseFormer \citep{niu2026phaseformer}. The $w$ phase tokens are embedded, exchange information through $M$ learnable routers (two cross-attentions), and are decoded by one shared linear predictor.}
\label{fig:phaseformer}
\Description{The folded input is embedded from n to d per phase. A group repeated N times contains an aggregate step (phases to M routers), a distribute step (routers back to phases) and a residual block with layer norm and MLP. A predictor from d to m and an unfold step with inverse RevIN produce the forecast.}
\end{figure}
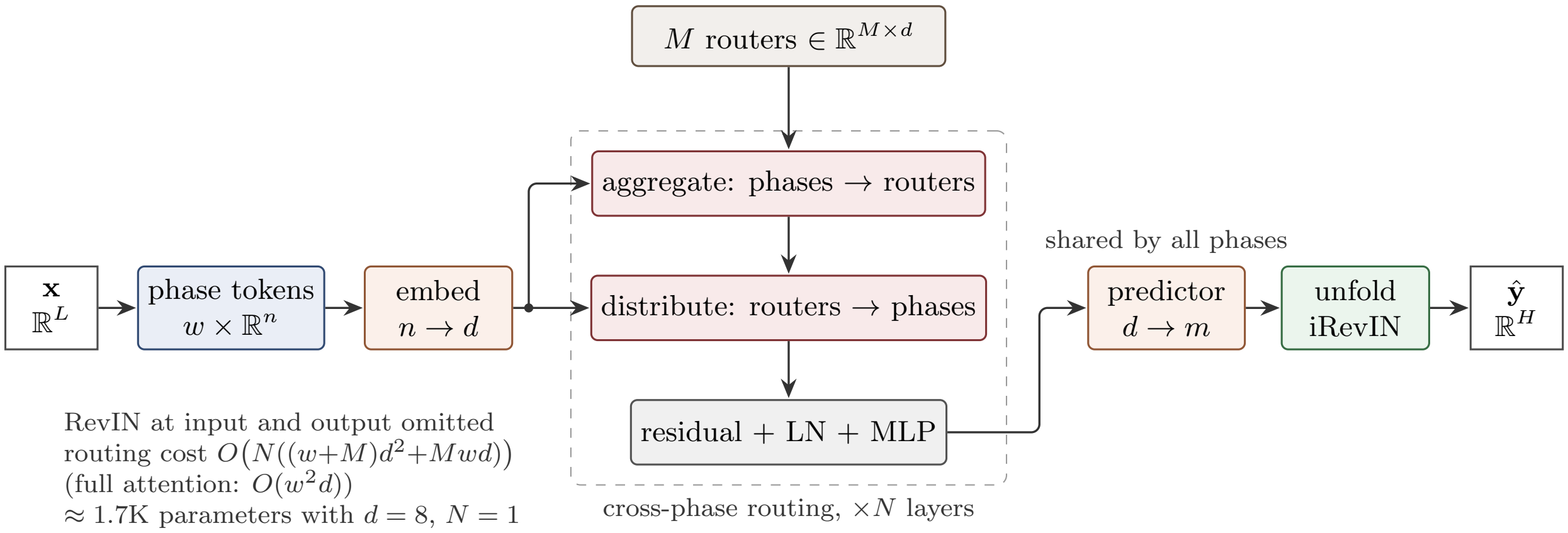

\paragraph{PhaseFormer.} \citet{niu2026phaseformer} (Figure~\ref{fig:phaseformer}) embed each phase
token into a low-dimensional space, exchange information across phases
through $M$ learnable routers, using one cross-attention from phases to
routers and one from routers to phases, and decode all phases with one shared linear
predictor. Unlike SparseTSF and TimeBase, it is non-linear and lets phases
interact. It reports the best average MSE on six of seven datasets at
$L=720$, with the largest gains on Traffic and Electricity
(Table~\ref{tab:acc}). Its parameter count depends strongly on the
configuration (Section~\ref{sec:audit}).

\paragraph{MoFo.} \citet{ma2025mofo} build period-structured 2D patches
(rows hold period-aligned steps) and add a period-aware modulator that
biases attention toward periodic patterns. It emphasizes memory efficiency
and training speed rather than minimal size.

\paragraph{Takeaway.} Folding turns periodicity into weight sharing, and a
basis compresses each phase vector further. The price is sensitivity to $w$, to multiple
periods, and to aperiodic data. Cross-phase interaction (PhaseFormer)
recovers accuracy on large datasets; its parameter cost there
cannot be pinned down from the released code (Section~\ref{sec:audit}).

\subsection{Cycle and Residual Models}
\label{sec:cycle}

\begin{table}[t]
\centering
\caption{Cycle and residual models. Parameters as in Table~\ref{tab:master} ($L=720$, $H=96$, ETTh1 configuration unless marked).}
\label{tab:fam-cycle}
\small
\begin{tabular}{llll}
\toprule
Model & Venue & Key idea & Params\\
\midrule
CycleNet \citep{lin2024cyclenet} & NeurIPS'24 & learnable cycle removed and restored & 69.4K\\
TQNet \citep{lin2025tqnet} & ICML'25 & periodic queries for channel attention & 3.02M\\
SSCNN \citep{deng2024sscnn} & NeurIPS'24 & sequential structured decomposition & $\sim$5K$^\S$\\
\bottomrule
\end{tabular}
\end{table}

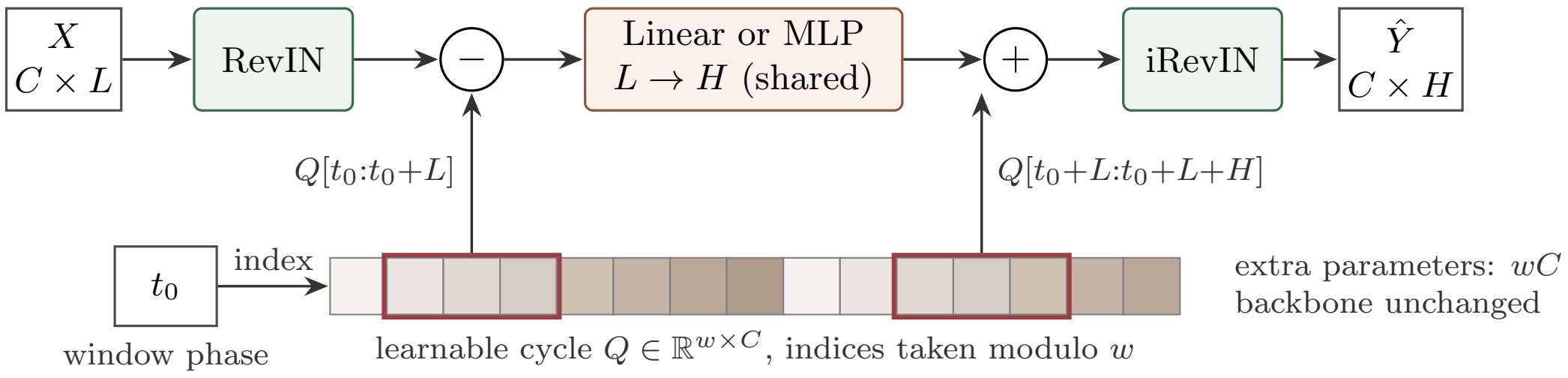
\begin{figure}[t]
\centering
\adjustbox{max width=\linewidth}{\begin{tikzpicture}[x=1cm,y=1cm]
\node[dat] (x) at (0,0) {$X$\\$C\times L$};
\node[blk=cNorm,minimum width=14mm] (n) at (1.85,0) {RevIN};
\node[op] (m) at (3.6,0) {$-$};
\node[blk=cLin,minimum width=28mm] (f) at (6.0,0) {Linear or MLP\\$L\to H$ (shared)};
\node[op] (p) at (8.4,0) {$+$};
\node[blk=cNorm,minimum width=14mm] (dn) at (10.05,0) {iRevIN};
\node[dat] (y) at (11.8,0) {$\hat Y$\\$C\times H$};
\foreach \a/\b in {x/n,n/m,m/f,f/p,p/dn,dn/y}\draw[arr] (\a)--(\b);
\foreach \c in {0,...,14}{\pgfmathtruncatemacro{\sh}{10+mod(\c,8)*9}
  \node[cell,minimum size=5mm,fill=cPar!\sh] (q\c) at (2.6+\c*0.5,-2.0) {};}
\node[draw=cAtt!80!black,line width=1.1pt,fit=(q1)(q3),inner sep=0.5pt] (s1) {};
\node[draw=cAtt!80!black,line width=1.1pt,fit=(q10)(q12),inner sep=0.5pt] (s2) {};
\draw[arr] (s1.north) -- node[pos=0.5,left,lbl] {$Q[t_0{:}t_0{+}L]$} (s1.north|-m.south);
\draw[arr] (s2.north) -- node[pos=0.5,right,lbl] {$Q[t_0{+}L{:}t_0{+}L{+}H]$} (s2.north|-p.south);
\node[lbl,anchor=north] at (q7.south) {learnable cycle $Q\in\R^{w\times C}$, indices taken modulo $w$};
\node[dat,minimum width=9mm,minimum height=7mm] (t0) at (0.9,-2.0) {$t_0$};
\draw[arr] (t0) -- node[above,lbl] {index} (q0.west);
\node[lbl,anchor=north] at (t0.south) {window phase};
\node[note,anchor=west] at (10.2,-2.0) {extra parameters: $wC$\\backbone unchanged};
\end{tikzpicture}}
\caption{CycleNet \citep{lin2024cyclenet}. A learnable cycle $Q\in\R^{w\times C}$ is indexed by the absolute phase $t_0$ of the window; the aligned cycle is removed before a linear or MLP backbone and the future cycle is added back.}
\label{fig:cyclenet}
\Description{Left-to-right chain: input of size C by L, RevIN, a minus node, a shared linear or MLP backbone from L to H, a plus node, inverse RevIN and the output. Below, a learnable w by C table is read at offset t0 and feeds both the minus node and the plus node. A note states that only wC parameters are added and the backbone is unchanged.}
\end{figure}

\paragraph{CycleNet.} \citet{lin2024cyclenet} (Figure~\ref{fig:cyclenet}) learn a recurrent cycle
$Q\in\R^{w\times C}$ shared by all windows (Residual Cycle Forecasting,
RCF). The cycle aligned with the window is subtracted from the input; a
linear layer or two-layer MLP forecasts the residual; the aligned future
cycle is added back. Alignment requires the absolute phase of each window
(\texttt{cycle\_index} in the code), so RCF uses information that CI models
without timestamps do not have. Parameters are those of the backbone plus
$wC$: 519{,}288 for CycleNet-Linear on ETTh1 at $L=H=720$ (from code). The
lightweight part is the cycle, not the backbone. RCF can also be added to other
backbones; the repository includes an iTransformer variant.

\begin{figure}[t]
\centering
\adjustbox{max width=\linewidth}{\begin{tikzpicture}[x=1cm,y=1cm]
\node[dat] (x) at (0,0) {$X$\\$C\times L$};
\node[blk=cNorm,minimum width=14mm] (n) at (1.75,0) {RevIN};
\node[blk=cAtt,minimum width=32mm,minimum height=12mm] (att) at (5.0,0) {multi-head attention\\over $C$ channel tokens\\{\scriptsize (embedding dim $=L$)}};
\node[blk=cPar,minimum width=40mm,minimum height=7mm] (tq) at (5.0,1.75) {temporal queries $\mathrm{TQ}[t_0{:}t_0{+}L]$, $\mathrm{TQ}\in\R^{w\times C}$};
\node[op] (p) at (7.4,0) {$+$};
\node[blk=cLin,minimum width=15mm] (i) at (8.95,0) {Linear\\$L\to d$};
\node[blk=cLin,minimum width=19mm] (mlp) at (11.0,0) {2-layer MLP\\+ residual};
\node[blk=cLin,minimum width=15mm] (o) at (13.05,0) {Linear\\$d\to H$};
\node[blk=cNorm,minimum width=14mm] (dn) at (14.9,0) {iRevIN};
\draw[arr] (x)--(n); \draw[arr] (n)--node[above,lbl]{K, V}(att); \draw[arr] (tq)--node[right,lbl]{Q}(att);
\draw[arr] (att)--(p); \draw[arr] (n.south) -- (1.75,-1.1) -| (p.south);
\foreach \a/\b in {p/i,i/mlp,mlp/o,o/dn}\draw[arr] (\a)--(\b);
\node[note,anchor=west] at (8.3,-1.3) {attention alone has $4L^2+4L$ parameters;\\0.66M in total at $L=96$, $d=512$};
\end{tikzpicture}}
\caption{TQNet \citep{lin2025tqnet}. Periodically shifted learnable vectors act as attention queries over channel tokens; keys and values are the normalized input. A shallow MLP then maps time to horizon.}
\label{fig:tqnet}
\Description{The input passes RevIN and enters a multi-head attention block whose query comes from a w by C table of learnable vectors and whose keys and values come from the channels. The result goes through a linear map from L to d, a two-layer MLP with residual connection, a linear map from d to H and inverse RevIN. A note states that attention alone has 4L squared plus 4L parameters and the model has 0.66M parameters at L = 96 and d = 512.}
\end{figure}
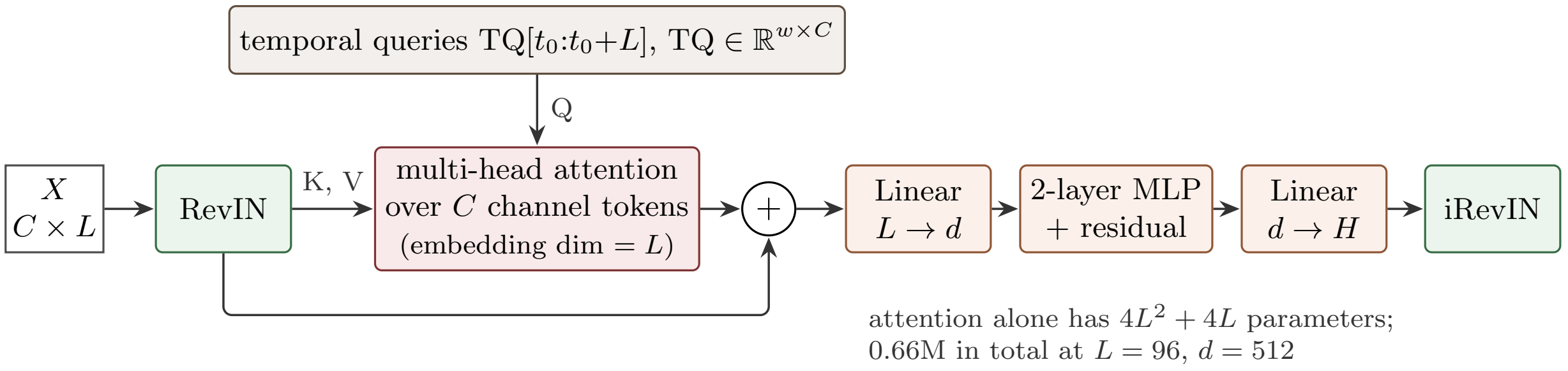

\paragraph{TQNet.} \citet{lin2025tqnet} (Figure~\ref{fig:tqnet}) use the periodically shifted
cycle vectors as \emph{queries} of a single multi-head attention over
channels, with keys and values from the input, followed by a shallow MLP.
This is the cheapest route to channel dependence in this family. However,
the attention embedding dimension equals $L$, so its size grows as $4L^2$.
With the default width $d=512$ we count $0.66$M parameters at $L=96$ and
$3.3$M at $L=720$ (from code). The default look-back is 96, and the authors'
README describes TQNet as about 1M parameters.

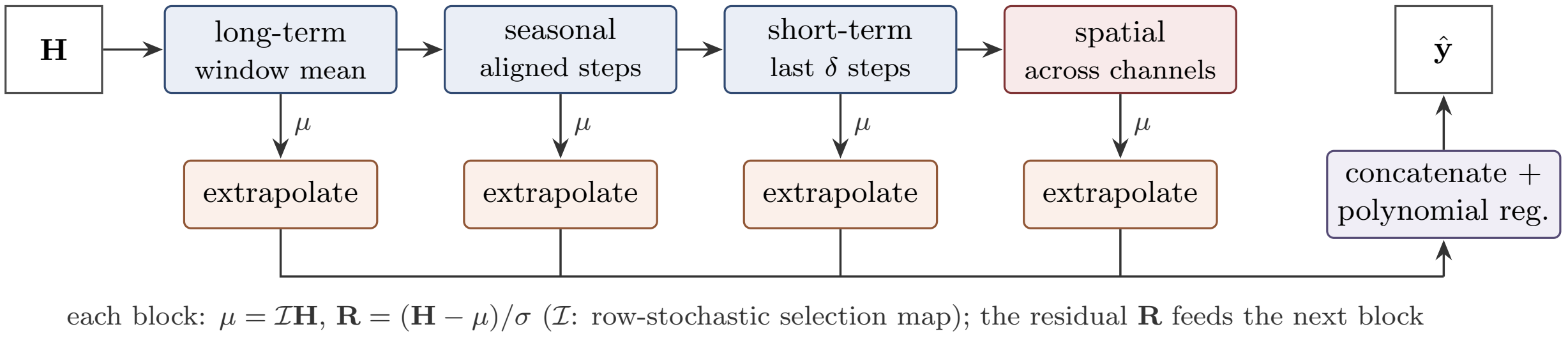
\begin{figure}[t]
\centering
\adjustbox{max width=\linewidth}{\begin{tikzpicture}[x=1cm,y=1cm]
\node[dat] (x) at (0,0) {$\mathbf H$};
\node[blk=cIn,minimum width=24mm] (lt) at (2.35,0) {long-term\\{\scriptsize window mean}};
\node[blk=cIn,minimum width=24mm] (se) at (5.25,0) {seasonal\\{\scriptsize aligned steps}};
\node[blk=cIn,minimum width=24mm] (st) at (8.15,0) {short-term\\{\scriptsize last $\delta$ steps}};
\node[blk=cAtt,minimum width=24mm] (sp) at (11.05,0) {spatial\\{\scriptsize across channels}};
\draw[arr] (x)--(lt); \draw[arr] (lt)--(se); \draw[arr] (se)--(st); \draw[arr] (st)--(sp);
\foreach \k in {lt,se,st,sp}{\node[blk=cLin,minimum width=20mm,minimum height=7mm] (e\k) at (\k|-0,-1.5) {extrapolate};\draw[arr] (\k)--node[right,lbl]{$\mu$}(e\k);}
\node[blk=cFreq,minimum width=24mm] (poly) at (14.4,-1.5) {concatenate +\\polynomial reg.};
\node[dat] (y) at (14.4,0) {$\hat{\mathbf y}$};
\draw[line width=0.65pt,black!80] (elt.south) -- (elt.south|-0,-2.35) -- (poly.south|-0,-2.35);
\foreach \k in {se,st,sp}\draw[line width=0.65pt,black!80] (e\k.south) -- (e\k.south|-0,-2.35);
\draw[arr] (poly.south|-0,-2.35) -- (poly.south);
\draw[arr] (poly)--(y);
\node[note,anchor=north west] at (0,-2.5) {each block: $\mu=\mathcal I\mathbf H$, $\mathbf R=(\mathbf H-\mu)/\sigma$ ($\mathcal I$: row-stochastic selection map); the residual $\mathbf R$ feeds the next block};
\end{tikzpicture}}
\caption{SSCNN \citep{deng2024sscnn}, simplified. Long-term, seasonal, short-term and spatial components are removed sequentially by attention-based normalization, extrapolated separately, and combined by a polynomial regression layer.}
\label{fig:sscnn}
\Description{A sequence of four blocks labelled window mean, aligned steps, last delta steps and across channels. Each block subtracts its estimated component and passes the residual to the next block; the component estimates are concatenated and combined by a polynomial regression into the forecast.}
\end{figure}

\paragraph{SSCNN.} \citet{deng2024sscnn} (Figure~\ref{fig:sscnn}) decompose each series into
long-term, seasonal, short-term and spatial components with
attention-based normalization, then extrapolate each. The seasonal
selection map acts only on period-aligned positions, which gives a
$\tfrac{1}{c^2}T_{\text{in}}T_{\text{out}}$ term in its parameter count, the
same $w^2$ reduction as Proposition~\ref{prop:kron}. It reports about 5K
parameters on ETT/Weather and 25K on Electricity/Traffic, with
$T_{\text{in}}=168$ for hourly data. It also reports higher FLOPs than
iTransformer because it keeps full-length representations
\citep{deng2024sscnn}. It is a useful reminder that few parameters do not
imply little computation.

\paragraph{Takeaway.} Cycle models store periodicity as parameters ($wC$)
instead of weight sharing. This is cheap per channel, adds
channel-specific seasonality, and opens a path to channel dependence.

\subsection{Light MLP and Mixer Models}
\label{sec:mlp}

\begin{table}[t]
\centering
\caption{Light MLP and mixer models. Parameters as in Table~\ref{tab:master} ($L=720$, $H=96$, ETTh1 configuration unless marked).}
\label{tab:fam-mlp}
\small
\begin{tabular}{llll}
\toprule
Model & Venue & Key idea & Params\\
\midrule
LightTS \citep{zhang2022lightts} & arXiv'22 & interval and continuous sampling MLPs & 182.8K\\
TSMixer \citep{chen2023tsmixer} & TMLR'23 & time and feature mixing MLPs & 1.56M\\
TiDE \citep{das2023tide} & TMLR'23 & dense encoder--decoder, covariates & 4.51M\\
TimeMixer \citep{wang2024timemixer} & ICLR'24 & multi-scale seasonal/trend mixing & 5.58M$^\S$\\
LiteMixer \citep{aityahia2025litemixer} & WINCOM'25 & entropy-gated enhancement, gated fusion & n.c.\\
\bottomrule
\end{tabular}
\end{table}

LightTS \citep{zhang2022lightts} uses continuous and interval sampling
followed by MLPs. TSMixer \citep{chen2023tsmixer} (Figure~\ref{fig:mixer}) alternates MLPs over time
and features; a different model with the same name was proposed by
\citet{ekambaram2023tsmixer}. TiDE \citep{das2023tide} is an MLP
encoder--decoder with covariates. With Time-Series-Library defaults these
are $10^5$--$10^7$ parameters (Table~\ref{tab:count}); they are light
relative to Transformers, not in the sense of this survey. Their advantage
is flexibility on aperiodic data and with covariates.

\begin{figure}[t]
\centering
\adjustbox{max width=\linewidth}{\begin{tikzpicture}[x=1cm,y=1cm]
\node[dat] (x) at (0,0) {$X$\\$C\times L$};
\node[blk=cIn,minimum width=19mm] (ds) at (2.0,0) {multi-scale\\downsampling};
\node[blk=cFreq,minimum width=19mm] (dft) at (4.35,0) {DFT\\season / trend};
\node[blk=gray,minimum width=19mm] (ent) at (6.7,0) {spectral\\entropy $>\tau$\,?};
\coordinate (br) at (8.0,0);
\node[blk=cAtt,minimum width=22mm] (lt) at (9.55,1.1) {LightTimesNet\\{\scriptsize kernels $\{3,5,7\}$}};
\node[circle,fill=black,inner sep=1.4pt] (mg) at (11.0,0) {};
\node[blk=cLin,minimum width=19mm] (mx) at (12.55,0) {multi-scale\\mixing};
\node[blk=cPar,minimum width=17mm] (gf) at (14.95,0) {gated fusion\\+ FFN};
\node[dat] (y) at (14.95,-1.75) {$\hat Y$\\$C\times H$};
\draw[arr] (x)--(ds); \draw[arr] (ds)--(dft); \draw[arr] (dft)--(ent);
\draw[line width=0.65pt,black!80] (ent.east)--(br);
\draw[arr] (br) |- (lt.west); \node[lbl,anchor=east] at (7.95,0.75) {yes};
\draw[arr] (br) |- (9.55,-1.0) -| (mg.south); \node[lbl,anchor=east] at (7.95,-0.6) {no};
\node[lbl,anchor=north] at (9.55,-1.02) {skip};
\draw[arr] (lt.east) -| (mg.north);
\draw[arr] (mg)--(mx); \draw[arr] (mx)--(gf); \draw[arr] (gf)--(y);
\node[lbl,anchor=north] at (12.55,-0.5) {seasonal bottom-up,\\trend top-down,\\shared MLPs};
\node[note,anchor=north west] at (0,-1.3) {$[g_s,g_t]=\sigma\big(\mathrm{Linear}([X_s,X_t])\big)$, \quad $X_{\text{fused}}=g_s\odot X_s+g_t\odot X_t$\\[2pt]
capacity per scale: $|\theta^{(i)}|=\alpha_i|\theta_{\text{base}}|$, \quad $\alpha_i=\log(s_i{+}1)/\sum_j\log(s_j{+}1)$};
\end{tikzpicture}}
\caption{LiteMixer \citep{aityahia2025litemixer}. Multi-scale series are split into season and trend by DFT; a LightTimesNet block runs only when the spectral entropy exceeds $\tau$; streams are mixed across scales with shared MLPs and combined by gated fusion.}
\label{fig:litemixer}
\Description{Left-to-right chain: input, multi-scale downsampling with factors 3, 5 and 7, a DFT box splitting season and trend, a spectral-entropy test that routes to an optional enhancement block, multi-scale mixing (seasonal bottom-up, trend top-down, shared MLPs), gated fusion with a feed-forward network, and a linear output head.}
\end{figure}
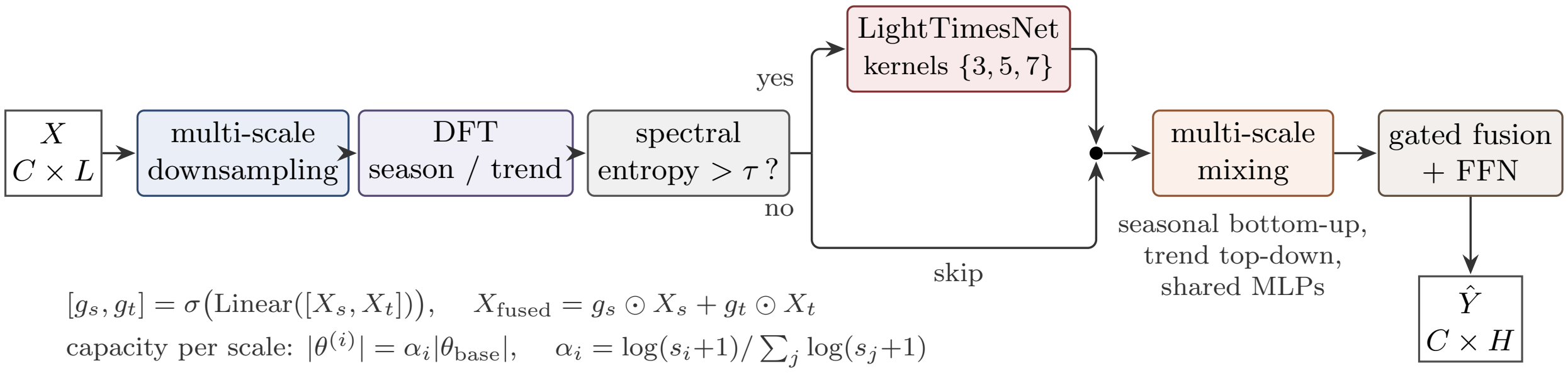

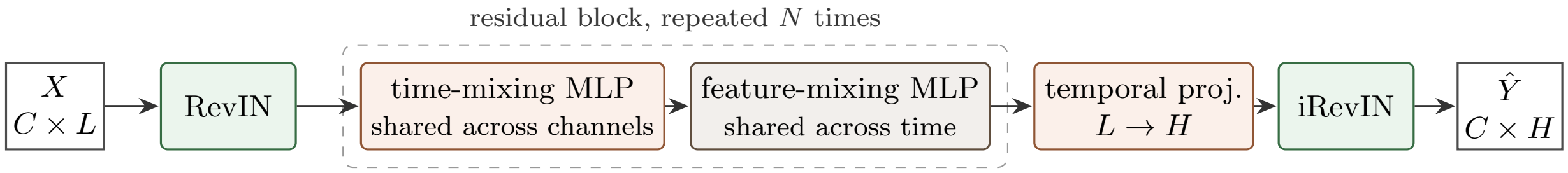
\begin{figure}[t]
\centering
\adjustbox{max width=\linewidth}{\begin{tikzpicture}[x=1cm,y=1cm]
\node[dat] (x) at (0,0) {$X$\\$C\times L$};
\node[blk=cNorm,minimum width=14mm] (n) at (1.8,0) {RevIN};
\node[blk=cLin,minimum width=30mm] (tm) at (4.75,0) {time-mixing MLP\\{\scriptsize shared across channels}};
\node[blk=cPar,minimum width=30mm] (fm) at (8.15,0) {feature-mixing MLP\\{\scriptsize shared across time}};
\node[grp,fit=(tm)(fm)] (g) {}; \node[lbl,anchor=south] at (g.north) {residual block, repeated $N$ times};
\node[blk=cLin,minimum width=20mm] (tp) at (11.3,0) {temporal proj.\\$L\to H$};
\node[blk=cNorm,minimum width=14mm] (dn) at (13.4,0) {iRevIN};
\node[dat] (y) at (15.1,0) {$\hat Y$\\$C\times H$};
\foreach \a/\b in {x/n,n/tm,tm/fm,fm/tp,tp/dn,dn/y}\draw[arr] (\a)--(\b);
\end{tikzpicture}}
\caption{TSMixer \citep{chen2023tsmixer}: residual blocks alternate an MLP over time (shared across channels) and an MLP over channels (shared across time), followed by a temporal projection. TiDE \citep{das2023tide} instead uses a dense encoder--decoder that also ingests covariates.}
\label{fig:mixer}
\Description{Left-to-right chain: input, RevIN, then a residual block repeated N times containing a time-mixing MLP shared across channels and a feature-mixing MLP shared across time, each with a residual connection. A temporal projection from L to H and inverse RevIN produce the output.}
\end{figure}

\paragraph{Multi-scale mixing.} TimeMixer \citep{wang2024timemixer} mixes
seasonal and trend components across several down-sampled scales with MLPs.
LiteMixer \citep{aityahia2025litemixer} (Figure~\ref{fig:litemixer})
keeps this decomposition but cuts its cost for edge and mobile deployment.
It poses the design as $\min_\theta|\theta|$ subject to
$\mathcal L(f_\theta(X),Y)\le\mathcal L_{\text{baseline}}$, with TimeMixer as the
baseline, and makes three changes.
\emph{Conditional enhancement}: a compressed TimesNet block (one shared FFT
period detection, kernels $\{3,5,7\}$, one convolution; 40\% of the TimesNet
block size) runs only when the spectral entropy of the segment exceeds a
threshold $\tau$, checked every $c$ steps, which skips it about half the time.
\emph{Gated fusion}: the seasonal and trend streams are combined as
$X_{\text{fused}}=g_s\odot X_s+g_t\odot X_t$ with
$[g_s,g_t]=\sigma(\mathrm{Linear}([X_s,X_t]))$, replacing MLP fusion and saving
over 80\% of the fusion parameters; one embedding is shared across scales.
\emph{Adaptive scale allocation}: scale $i$ receives a share
$\alpha_i=\log(s_i+1)/\sum_j\log(s_j+1)$ of the capacity, and the mixing MLPs
are shared across scales. Against TimeMixer, LiteMixer uses 15.75\% fewer
parameters on average (up to 32\% at $H=96$ on ETT) and reaches the lowest
average MSE on four of five benchmarks (Weather, ETTh1, ETTm1, ETTm2), and
second on ETTh2 \citep{aityahia2025litemixer}. In the terms of this survey it
is a size reduction of a TimeMixer-scale model, not a new structural
constraint; it is therefore complementary to the folding
family.

\subsection{Other Lightweight Designs: a Residual Group}
\label{sec:other}

\begin{table}[t]
\centering
\caption{Other designs. Parameters as in Table~\ref{tab:master} ($L=720$, $H=96$, ETTh1 configuration unless marked).}
\label{tab:fam-other}
\small
\begin{tabular}{llll}
\toprule
Model & Venue & Key idea & Params\\
\midrule
N-BEATS \citep{oreshkin2020nbeats} & ICLR'20 & basis-expansion residual MLP & n.c.\\
PITS \citep{lee2024pits} & ICLR'24 & patch-independent MLP embedding & 1.10M$^\ast$\\
xPatch \citep{stitsyuk2025xpatch} & AAAI'25 & EMA split, linear and CNN streams & 144.0K$^\ast$\\
SOFTS \citep{han2024softs} & NeurIPS'24 & series--core channel fusion (STAR) & 510.8K$^\ast$\\
SegRNN \citep{lin2023segrnn} & arXiv'23 & segment recurrence, parallel decoding & 1.63M\\
TimeKAN \citep{huang2025timekan} & ICLR'25 & multi-band KAN layers & 12.8K$^\ast$\\
S-Mamba \citep{wang2024smamba} & arXiv'24 & bidirectional Mamba over variates & 1.15M$^\ast$\\
\bottomrule
\end{tabular}
\end{table}

This section is a residual group, not a family. It collects efficient
models that are standard baselines in the lightweight literature but are not
built around a linear, spectral, folding or cycle constraint: patch models,
recurrence, Kolmogorov--Arnold networks, channel mixing, state-space models,
convolution and basis expansion. They share no structural property beyond
that, so we discuss them one by one. Table~\ref{tab:beyond} gives the linear
component of each.

\paragraph{Patch MLPs.} PITS \citep{lee2024pits} embeds each patch
independently with an MLP, without attention across patches, and shows that
patch-independent embeddings are competitive with PatchTST. With its
released ETTh1 configuration ($L=512$, $d=256$, patch 24) it has 1.10M
parameters (from code). xPatch \citep{stitsyuk2025xpatch} splits the series with
an exponential moving average into trend and seasonal streams, forecasts the
trend with a linear stream and the seasonal part with a patch-based
convolutional stream; its unified configuration ($L=96$) has 144.0K
parameters (from code).

\paragraph{Recurrence.} SegRNN \citep{lin2023segrnn} revives RNNs for long
horizons with two changes: the recurrence runs over segments instead of
points, and all horizon segments are decoded in parallel (parallel
multi-step forecasting). This cuts the number of recurrent steps from $L+H$
to $L/s+1$ for segment length $s$ and, per its authors, reduces runtime and
memory by more than 78\% relative to Transformer baselines. With $d=512$ it
has 1.63M parameters at $L=720$ (from code); the cost is in the hidden state,
not in the sequence length.

\paragraph{Kolmogorov--Arnold networks.} TimeKAN \citep{huang2025timekan}
decomposes the series into frequency bands with cascaded down-sampling,
learns each band with multi-order KAN layers (Chebyshev bases), and mixes
the bands back. Its released ETTh1 configuration ($d=16$) has 12.8K
parameters (from code), which places it at the FITS scale with a non-linear
model.

\paragraph{Efficient channel mixing.} SOFTS \citep{han2024softs} addresses the
main weakness of channel-independent models at linear cost. Its STAR module
aggregates all series into one global \emph{core} representation, then
dispatches it back and fuses it with each series, so channel interaction
costs $O(C)$ instead of the $O(C^2)$ of attention over variates. Its ETTh1
configuration has 511K parameters (from code).

\paragraph{State-space models.} Selective state-space models such as Mamba
scale linearly with sequence length. S-Mamba \citep{wang2024smamba}
embeds each variate with a linear layer, applies a bidirectional Mamba
layer across variates to capture their correlations, and a feed-forward
network for temporal dependence. With its released ETTh1 configuration
($d=256$, state size 2) it has 1.15M parameters (from code, counted with a
shape-level stand-in for the Mamba block). Its efficiency is relative to
variate-token Transformers such as iTransformer, not to the folding family.

\paragraph{Basis-expansion MLPs.} N-BEATS \citep{oreshkin2020nbeats} is an
early deep forecaster built only from fully connected blocks with backward
and forward residual links. Its interpretable variant constrains the block
outputs to polynomial (trend) and Fourier (seasonality) bases, the same
idea as the fixed-basis models of Section~\ref{sec:frequency}. N-HiTS
extends it with multi-rate sampling and hierarchical interpolation.

\paragraph{Other efficient backbones.} ModernTCN \citep{luo2024moderntcn}
uses large depth-wise convolutions, PDF \citep{dai2024pdf} decouples short-
and long-term variation along detected periods, N-HiTS
\citep{challu2023nhits} is discussed above, and Koopa
\citep{liu2023koopa} uses Koopman predictors for non-stationary dynamics.
They target efficiency relative to Transformers rather than minimal size.

\section{Lightweight Pre-trained Models}
\label{sec:pretrained}

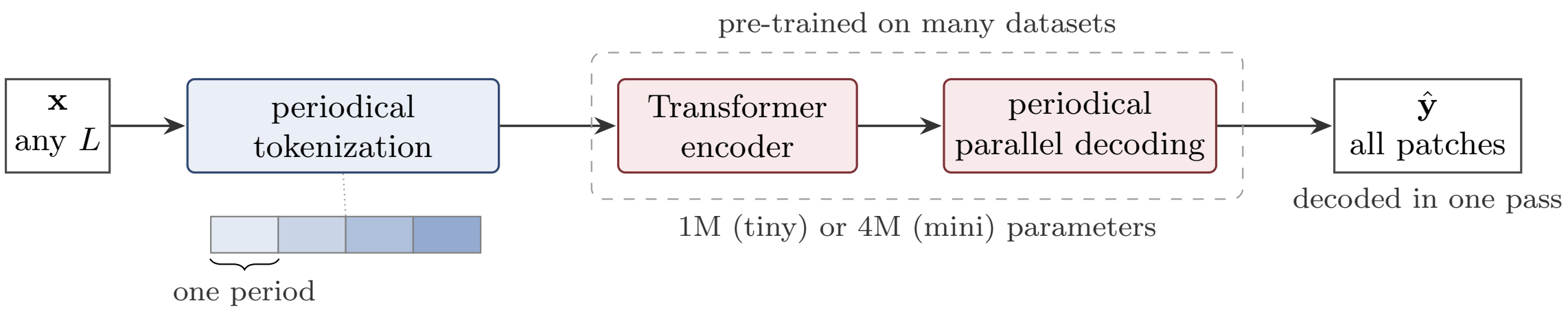
\begin{figure}[t]
\centering
\adjustbox{max width=\linewidth}{\begin{tikzpicture}[x=1cm,y=1cm]
\node[dat] (x) at (0,0) {$\mathbf x$\\any $L$};
\node[blk=cIn,minimum width=30mm] (tok) at (2.75,0) {periodical\\tokenization};
\node[blk=cAtt,minimum width=23mm] (enc) at (6.55,0) {Transformer\\encoder};
\node[blk=cAtt,minimum width=23mm] (dec) at (9.85,0) {periodical\\parallel decoding};
\node[dat,minimum width=18mm] (y) at (13.2,0) {$\hat{\mathbf y}$\\all patches};
\draw[arr] (x)--(tok); \draw[arr] (tok)--(enc); \draw[arr] (enc)--(dec); \draw[arr] (dec)--(y);
\node[grp,fit=(enc)(dec),inner sep=7pt] (g) {};
\node[lbl,anchor=south] at (g.north) {pre-trained on many datasets};
\node[lbl,anchor=north] at (g.south) {1M (tiny) or 4M (mini) parameters};
\foreach \k in {0,...,3}{\pgfmathtruncatemacro{\sh}{15+\k*15}
  \node[cell,minimum width=6.5mm,minimum height=3.5mm,fill=cIn!\sh] (p\k) at (1.8+\k*0.65,-1.05) {};}
\draw[decorate,decoration={brace,amplitude=3pt,mirror}] ($(p0.south west)+(0,-0.05)$) -- ($(p0.south east)+(0,-0.05)$) node[midway,below=2pt,lbl] {one period};
\draw[black!40,densely dotted] (tok.south) -- ($(p1.north)!0.5!(p2.north)$);
\node[lbl,anchor=north] at (y.south) {decoded in one pass};
\end{tikzpicture}}
\caption{LightGTS \citep{wang2025lightgts}. Tokens are cut at the period length of each dataset, so one pre-trained encoder--decoder can serve series with different sampling rates; horizon patches are decoded in parallel.}
\label{fig:lightgts}
\Description{Left-to-right chain: an input of any length, periodical tokenization, a Transformer encoder, periodical parallel decoding and the forecast. A note gives the sizes of the two released models, 1M parameters (tiny) and 4M parameters (mini).}
\end{figure}

Foundation models target zero-shot forecasting. TimesFM has about 200M
parameters \citep{das2024timesfm}, and Chronos tokenizes values by scaling and
quantization and trains T5 models from 20M to 710M parameters
\citep{ansari2024chronos}; Moirai comes in 14M, 91M and 311M sizes
\citep{woo2024moirai}. A related line adapts large language models to
forecasting. \citet{tan2024llm} show that removing the language model from
three such methods, or replacing it with one attention layer, does not
degrade accuracy and often improves it, at orders of magnitude less
compute; their patch-and-attention baseline (PAttn) is a useful small
reference. Two works show that pre-training can be
far smaller. Tiny Time Mixers \citep{ekambaram2024ttm} start from 1M parameters and
build on the IBM TSMixer \citep{ekambaram2023tsmixer}. To pre-train such a
small model on data with very different sampling rates, TTM adds adaptive
patching across layers, diverse resolution sampling, and a resolution
prefix token. It is pre-trained channel-independently and fine-tuned with
channel mixing and exogenous inputs, and it runs on CPU-only machines. LightGTS \citep{wang2025lightgts} (Figure~\ref{fig:lightgts})
introduces periodical tokenization and periodical parallel decoding and
reports two sizes, LightGTS-tiny (1M) and LightGTS-mini (4M). The
periodicity prior of Section~\ref{sec:folding} thus also transfers to
pre-training. In a related direction, SparseTSF reports that a model
trained on ETTh2 or Electricity transfers to ETTh1 with little loss when
the periods match \citep[Table~7]{lin2024sparsetsf}.

\section{Where Lightweight Forecasters Matter}
\label{sec:apps}

The standard benchmarks of Table~\ref{tab:data} come from four application
domains. Their properties (sampling rate, number of channels, strength of the
period) are exactly the quantities that decide which family of
Section~\ref{sec:families} fits, and they explain why size matters in each
domain.

\paragraph{Energy systems.} ETT records the oil temperature and loads of
electricity transformers \citep{zhou2021informer}, and Electricity records the
hourly consumption of 321 clients \citep{wu2021autoformer}. Both have a strong
daily period, and Electricity also a weekly one. Forecasts are often needed for
many meters or substations at once, and may have to run on or near the
device, so per-series cost multiplies with the number of series. This is the regime
where channel-independent folding models are most attractive: one predictor
of a few hundred parameters serves every meter, and the closed-form results
of Section~\ref{sec:closedform} show that on hourly ETT and
Electricity data the phase-shared constraint costs little or nothing at long
horizons.

\paragraph{Transportation.} Traffic records hourly road occupancy from 862
sensors \citep{wu2021autoformer}. It combines a daily and a weekly cycle and
many correlated channels. Section~\ref{sec:audit} shows that this is where
the purely linear folding models fall behind and where cross-phase
interaction (PhaseFormer) pays off; cross-channel designs
(CycleNet, TQNet, SOFTS) target the same gap. The size of the PhaseFormer
configuration behind that result is unresolved (Section~\ref{sec:audit}).
The fold itself is not the problem: on Traffic, the phase-shared map costs
only 0.5--2\% against the full linear map (Section~\ref{sec:closedform}), so
the gap lies between linear and non-linear models.

\paragraph{Weather and environment.} Weather records 21 meteorological
variables every 10 minutes \citep{wu2021autoformer}. At this resolution the
daily period spans 144 steps, longer than the look-back used by many models,
and the series is smooth over short windows. A daily fold is then a poor
constraint, costing 9\% at $H=720$ and more at short horizons
in our closed-form fits (Section~\ref{sec:closedform}); small folding
periods, band-limited maps, or MLP models are safer choices.

\paragraph{Edge and embedded deployment.} Several lightweight models are
motivated explicitly by constrained hardware: SparseTSF targets devices with
very limited computational resources \citep{lin2024sparsetsf}, LiteMixer
targets edge, mobile and wireless settings \citep{aityahia2025litemixer}, and
TTM runs on CPU-only machines \citep{ekambaram2024ttm}. In this regime the
relevant budget is not only parameters but also latency, memory and energy
per forecast (Section~\ref{sec:deploy}); energy per forecast is rarely reported, which we list as an open problem
(Section~\ref{sec:open}).

\paragraph{Takeaway.} The benchmark domains differ in the two properties
that matter most for lightweight design: how stable and dominant the period
is, and how many channels share it. Reporting results per domain, rather
than as a single average, would make the trade-offs of each family visible.

\section{Lessons from a Code-Level Audit}
\label{sec:audit}

\paragraph{Which repositories are audited.} The audit is a purposive sample,
not a census. It contains two groups of third-party code. The first is the
official code of the linear, frequency, folding and cycle models whose sizes
this literature compares: DLinear and NLinear, RLinear, FITS, FreTS,
Amplifier, SparseTSF, TimeBase, PhaseFormer, CycleNet and TQNet. The second
is one or two efficiency-claiming models from the other designs: SegRNN,
TimeKAN, SOFTS, xPatch, PITS and S-Mamba. Reference models are counted from
the Time-Series-Library. The audit excludes LiteMixer and RAYQUAZA,
pre-trained models, whose size is fixed by the released weights,
and seven models whose code we did not run: FilterNet, MoFo, PDF, SSCNN,
ModernTCN, Koopa and N-BEATS. Their sizes in Table~\ref{tab:master} are
reported by their authors or marked n.c. Repositories and commits are listed
in Appendix~\ref{app:code}.

\subsection{Parameter counts from the official code}

Table~\ref{tab:count} gives parameter counts obtained by instantiating each
model's constructor with its released configuration
(Appendix~\ref{app:counting}). Where the configuration is known, the counts
reproduce the published ones exactly: SparseTSF 41 and 145 at $L=96$
\citep[Table~4]{lin2024sparsetsf} and 925 (``0.92K'') at $L=H=720$
\citep[Table~3]{lin2024sparsetsf}, TimeBase 396
\citep{huang2025timebase}, TimeBase 214 at $H=96$
\citep[Table~4]{niu2026phaseformer}, PhaseFormer 1{,}716 for the
``$\approx$1.72K'' variant \citep[A.4.1]{niu2026phaseformer}, DLinear
485{,}280 at $L=336$ and 1{,}038{,}240 at $L=720$
\citep{lin2024sparsetsf,huang2025timebase}.

\begin{table}[t]
\centering
\caption{Trainable parameters computed from official code. ETTh1
configuration ($C=7$, $w=24$) unless noted. Reference models use
Time-Series-Library defaults ($d_{\text{model}}=512$).}
\label{tab:count}
\small
\begin{tabular}{lrrrr}
\toprule
Model (repository config) & $96\to96$ & $336\to96$ & $720\to96$ & $720\to720$\\
\midrule
SparseTSF & 41 & 81 & 145 & 925\\
TimeBase ($R=6$) & 58 & 118 & 214 & 396\\
PhaseFormer ($d=8$, 1 layer, 8 routers) & 1{,}508 & 1{,}588 & 1{,}716 & 1{,}950\\
FITS ($h=2$, real count) & 1{,}680 & 4{,}182 & 11{,}826 & 21{,}024\\
Linear & 9{,}312 & 32{,}352 & 69{,}216 & 519{,}120\\
CycleNet-Linear & 9{,}480 & 32{,}520 & 69{,}384 & 519{,}288\\
DLinear & 18{,}624 & 64{,}704 & 138{,}432 & 1{,}038{,}240\\
\midrule
LightTS & 121{,}002 & 144{,}442 & 182{,}778 & 673{,}242\\
TSMixer \citep{chen2023tsmixer} & 222{,}510 & 737{,}550 & 1{,}561{,}614 & 2{,}011{,}518\\
TQNet ($d=512$) & 661{,}640 & 1{,}200{,}200 & 3{,}020{,}360 & 3{,}340{,}472\\
TiDE & 2{,}534{,}141 & 3{,}294{,}461 & 4{,}510{,}973 & 10{,}729{,}133\\
FreTS & 3{,}236{,}832 & 11{,}101{,}152 & 23{,}684{,}064 & 23{,}844{,}432\\
iTransformer & 6{,}404{,}704 & 6{,}527{,}584 & 6{,}724{,}192 & 7{,}044{,}304\\
PatchTST & 6{,}903{,}904 & 8{,}378{,}464 & 10{,}737{,}760 & 39{,}492{,}304\\
\bottomrule
\end{tabular}
\end{table}

\subsection{The same baseline, reported differently}

Table~\ref{tab:inconsistent} collects parameter counts that different papers
report for the same baselines. Differences reach 289$\times$ for FITS (462
against 133.6K parameters on Electricity, at horizons that explain less than
$2\times$ by the count from code) and 31$\times$ for SparseTSF (145 from code against
4.509K reported). Some are explained by the horizon, the cutoff frequency,
per-channel (\texttt{individual}) weights, or the complex-weight
convention; the papers do not state which. Since these counts are the
headline argument of this literature, the lack of a common convention is a
real problem.

\begin{table}[t]
\centering
\caption{Reported parameter counts for the same models. All on
Electricity with $L=720$ unless noted. ``From code'' is computed from the
official code with the repository defaults ($C$-independent models).}
\label{tab:inconsistent}
\small
\begin{tabular}{lllll}
\toprule
Model & Source & Setting & Reported & From code\\
\midrule
FITS & \citet[Tab.~3]{lin2024sparsetsf} & $H=720$ & 10.5K & 21.0K ($h=2$)\\
FITS & \citet[Tab.~3]{huang2025timebase} & $H=720$ & 133.60K & 21.0K ($h=2$)\\
FITS & \citet[Tab.~4]{niu2026phaseformer} & $H=96$ & 462 & 11.8K ($h=2$)\\
SparseTSF & \citet[Tab.~3]{lin2024sparsetsf} & $H=720$ & 0.92K & 925\\
SparseTSF & \citet[Tab.~3]{huang2025timebase} & $H=720$ & 1.00K & 925\\
SparseTSF & \citet[Tab.~4]{niu2026phaseformer} & $H=96$ & 4.509K & 145\\
SparseTSF & \citet[Tab.~4]{niu2026phaseformer} & Traffic, $H=96$ & 17.949K & 145\\
TimeBase & \citet[Tab.~4]{niu2026phaseformer} & $H=96$ & 214 & 214\\
PatchTST & \citet[Tab.~3]{lin2024sparsetsf} & $H=720$, own $L$ & 6.31M & config-dependent\\
PatchTST & \citet[Tab.~3]{huang2025timebase} & $H=720$ & 8.69M & config-dependent\\
PatchTST & \citet[Tab.~4]{niu2026phaseformer} & $H=96$ & 1.373M & config-dependent\\
\bottomrule
\end{tabular}
\end{table}

\paragraph{PhaseFormer configuration.} PhaseFormer's Table~4 reports
1.156K parameters on six of seven datasets (308 on Weather), and its appendix describes a
two-layer, four-router model for Traffic and Electricity. The released
Traffic script sets latent width $32$ with two layers for $H=96$ and width
$128$ with one layer for $H\ge192$. Counted from code, these give about 38.7K and
276K parameters at $L=720$. The paper's own ablation reports a 37K variant
with similar Traffic MSE to the 1.7K variant at $H=96$ (0.360 vs.\ 0.361)
\citep[Table~7]{niu2026phaseformer}. We could not determine which
configuration produced the Traffic numbers of its Table~1. We report this
as a reproducibility gap, not as an error.

\subsection{Evaluation code}

A bug inherited from the Informer code base sets \texttt{drop\_last=True}
in the test loader, which silently discards the last incomplete batch;
results then depend on the batch size, mostly on small datasets such as
ETTh1/ETTh2. It was documented in the FITS repository and by the TFB
benchmark \citep{qiu2024tfb}. Table~\ref{tab:droplast} shows the state of the test
loaders in the current repositories. Several papers publish pre-fix and
post-fix tables side by side (e.g., SparseTSF, Appendix~D.1); when
comparing across papers, the post-fix tables must be used. Besides RTSF, which predates the bug report, five audited repositories still drop the last test batch by default: FreTS, Amplifier, PITS, TimeKAN and xPatch. S-Mamba also sets \texttt{drop\_last=True} but tests with batch size 1, so no window is lost. xPatch
documents this explicitly: its default setting keeps the ``drop-last trick''
to match published baselines, and its README points to separate results
without it from the TFB/OpenTS benchmark \citep{qiu2024tfb}. Numbers from
these default settings should not be compared directly with post-fix tables.

\begin{table}[t]
\centering
\caption{Test data loader in the current official repositories
(checked September 2026; last-commit dates in parentheses).}
\label{tab:droplast}
\footnotesize
\begin{tabular}{ll}
\toprule
Repository & Test loader\\
\midrule
LTSF-Linear (DLinear; 2024-01) & \texttt{drop\_last=False}\\
RTSF (RLinear; 2023-09) & \texttt{drop\_last=True} (predates the bug report)\\
FITS (2025-06) & \texttt{drop\_last=False}\\
SparseTSF (2026-03) & \texttt{drop\_last=False}\\
CycleNet (2026-01), TQNet (2026-05) & \texttt{drop\_last=False}\\
TimeBase (2025-09) & \texttt{drop\_last=False}\\
PhaseFormer (2026-03), Time-Series-Library & \texttt{drop\_last=False}\\
SegRNN (2026-02), SOFTS (2025-07) & \texttt{drop\_last=False}\\
S-Mamba (2025-05) & \texttt{drop\_last=True}, test batch size 1 (no window lost)\\
FreTS (2024-03) & \texttt{drop\_last=True}\\
Amplifier (2026-03) & \texttt{drop\_last=True}\\
PITS (2024-05) & \texttt{drop\_last=True}\\
TimeKAN (2025-02) & \texttt{drop\_last=True}\\
xPatch (2025-04) & \texttt{drop\_last=True} (fair results reported separately)\\
\bottomrule
\end{tabular}
\end{table}

\subsection{Accuracy under a matched look-back}

Table~\ref{tab:acc} reproduces the one comparison we found that runs the
lightweight models and strong Transformers at the same look-back ($L=720$)
in one paper \citep[Table~1]{niu2026phaseformer}. Baselines were run from
their official code by the PhaseFormer authors. Three
patterns stand out. On the ETT datasets, the three folding models are within
$0.016$ of each other and match or beat PatchTST to within $0.008$. On
Electricity, the linear folding models (TimeBase, SparseTSF) match PatchTST
(0.167 and 0.168 vs.\ 0.169) but trail the two models with cross-phase or
cross-channel interaction, PhaseFormer (0.160) and iTransformer (0.165). On
Traffic, they fall behind PatchTST (0.418 and 0.413 vs.\ 0.394), and
cross-phase interaction (PhaseFormer, 0.386) closes the gap. This is
consistent with the folding view: Traffic combines a daily and a weekly
period with many correlated channels, which no linear map absorbs: in
Section~\ref{sec:closedform}, the phase-shared constraint costs only
0.5--2\% against the full linear map on Traffic, so the gap lies in
linearity, not in folding. On Electricity the gain from interaction is smaller, and the
evidence comes from one paper.

\begin{table}[t]
\centering
\caption{MSE averaged over $H\in\{96,192,336,720\}$, $L=720$ for all
models, as reported by \citet[Table~1]{niu2026phaseformer}. Bold: best in
row. See Section~\ref{sec:audit} for the PhaseFormer configuration caveat.}
\label{tab:acc}
\small
\begin{tabular}{lcccccc}
\toprule
Dataset & PhaseFormer & TimeBase & SparseTSF & FITS & PatchTST & iTransformer\\
\midrule
ETTh1 & \textbf{0.403} & 0.404 & 0.406 & 0.419 & 0.420 & 0.453\\
ETTh2 & 0.346 & 0.347 & 0.345 & \textbf{0.334} & 0.344 & 0.392\\
ETTm1 & \textbf{0.346} & 0.356 & 0.362 & 0.359 & 0.354 & 0.370\\
ETTm2 & \textbf{0.250} & \textbf{0.250} & 0.252 & 0.285 & 0.251 & 0.278\\
Weather & \textbf{0.223} & 0.227 & 0.243 & 0.241 & \textbf{0.223} & 0.233\\
Electricity & \textbf{0.160} & 0.167 & 0.168 & 0.172 & 0.169 & 0.165\\
Traffic & \textbf{0.386} & 0.418 & 0.413 & 0.410 & 0.394 & 0.406\\
\bottomrule
\end{tabular}
\end{table}

\begin{figure}[t]
\centering
\includegraphics[width=\linewidth]{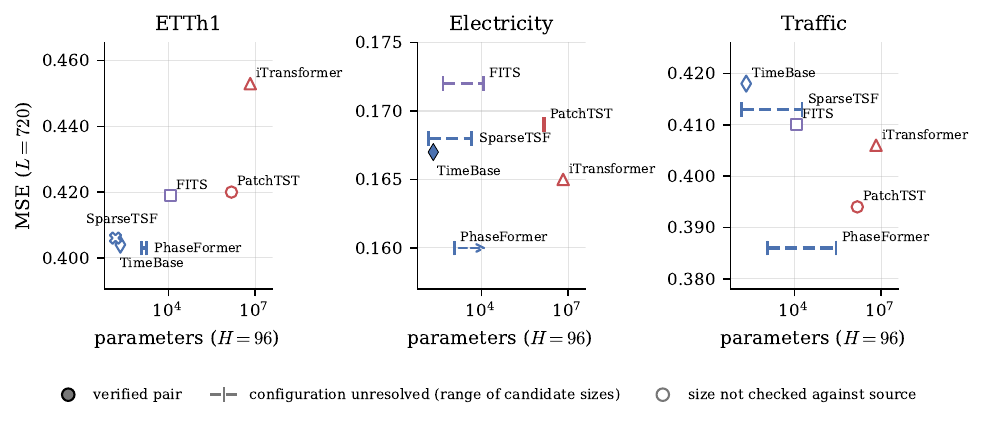}
\caption{Accuracy versus size for the models of Table~\ref{tab:acc}, each accuracy paired only with sizes we could tie to it. MSE from \citet[Table~1]{niu2026phaseformer} ($L=720$, averaged over $H$); sizes at $H=96$, the setting at which \citet[Table~4]{niu2026phaseformer} reports them. Filled marker: verified pair, the count from the official code equals the size reported by the accuracy source. Dashed range: configuration unresolved, spanning the size reported by the source and the count from code; for PhaseFormer on Traffic it also covers the released script (38.7K at $H=96$, 276K for $H\ge192$), and on Electricity its appendix describes a larger configuration of unknown size (open arrow). Hollow marker: the count from code for the repository configuration (for PatchTST $d_{\text{model}}=128$, for iTransformer the library default), not checked against the source.}
\label{fig:paretolit}
\Description{Scatter plot of test mean squared error against parameter count on a log scale, one marker per model and dataset.}
\end{figure}

Figure~\ref{fig:paretolit} places the same numbers against
model size and separates verified pairs from unresolved ones. Only one pair is
verified: TimeBase on Electricity, where the count from code equals the 214
parameters reported by the accuracy source. It sits nearly four orders of
magnitude to the left of PatchTST (1.4--1.5M) at slightly better accuracy
(0.167 vs.\ 0.169). For SparseTSF, FITS and PatchTST, the source reports
sizes that differ from the repository configuration (by up to $124\times$ for
SparseTSF on Traffic), and for PhaseFormer on Electricity and Traffic the
configuration behind the accuracy is unknown (Section~\ref{sec:audit}). Even
at the upper end of each candidate range, the folding and frequency models
remain at least $75\times$ smaller than the Transformers, so that ordering is
robust. PhaseFormer's size on Traffic is not determined within a factor of
about 240, and we draw no accuracy--size trade-off from it.

\subsection{Look-back sensitivity}

\begin{figure}[t]
\centering
\includegraphics[width=0.85\linewidth]{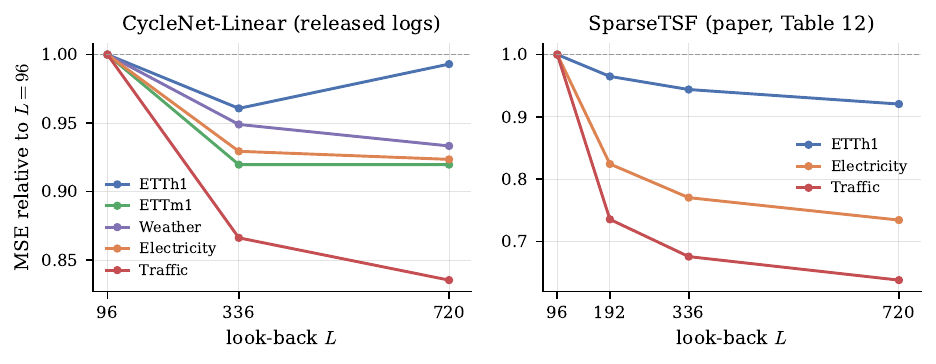}
\caption{MSE relative to $L=96$ as a function of the look-back (data of Table~\ref{tab:lookback}).}
\label{fig:lookback}
\Description{Line plot of mean squared error relative to its value at a look-back of 96, as a function of the look-back length, one line per model.}
\end{figure}

Table~\ref{tab:lookback} reports average MSE of CycleNet-Linear computed
from the per-seed results released in its repository (five seeds, our
averaging), and of SparseTSF from its paper. The effect of $L$ is often
larger than the gap between competing models in Table~\ref{tab:acc}. On
Traffic, moving from $L=96$ to $L=720$ lowers CycleNet's MSE from 0.486 to
0.406; the weekly period ($168$ steps) is not covered by $L=96$, as
SparseTSF also notes \citep[D.2]{lin2024sparsetsf}. The optimum is not
always the longest window: on ETTh1, CycleNet is best at $L=336$.
Comparisons at a single fixed $L$ can therefore reverse rankings. Figure~\ref{fig:lookback} plots the same data
relative to $L=96$. The two models come from different evaluation protocols
(CycleNet post-fix, SparseTSF pre-fix; Section~\ref{sec:audit}), so the table
shows the effect of $L$ within each model and should not be read across
models.

\begin{table}[t]
\centering
\caption{Average MSE over $H\in\{96,\dots,720\}$ versus look-back $L$.
CycleNet-Linear: our average of the released per-seed logs (post-fix
protocol). SparseTSF: \citet[Table~12]{lin2024sparsetsf} (pre-fix protocol).
Bold: best $L$ within each row. Compare within a block only: the two blocks
use different test loaders (Table~\ref{tab:droplast}).}
\label{tab:lookback}
\small
\begin{tabular}{llcccc}
\toprule
Model & Dataset & $L=96$ & $L=192$ & $L=336$ & $L=720$\\
\midrule
CycleNet-Linear & ETTh1 & 0.433 & -- & \textbf{0.416} & 0.430\\
CycleNet-Linear & ETTm1 & 0.386 & -- & \textbf{0.355} & \textbf{0.355}\\
CycleNet-Linear & Weather & 0.255 & -- & 0.242 & \textbf{0.238}\\
CycleNet-Linear & Electricity & 0.170 & -- & 0.158 & \textbf{0.157}\\
CycleNet-Linear & Traffic & 0.486 & -- & 0.421 & \textbf{0.406}\\
\midrule
SparseTSF & ETTh1 & 0.428 & 0.413 & 0.404 & \textbf{0.394}\\
SparseTSF & Electricity & 0.222 & 0.183 & 0.171 & \textbf{0.163}\\
SparseTSF & Traffic & 0.635 & 0.467 & 0.429 & \textbf{0.405}\\
\bottomrule
\end{tabular}
\end{table}

\section{Controlled Experiments}
\label{sec:closedform}

Published comparisons mix architectures, optimizers, and tuning budgets.
To isolate the effect of the structural constraint alone, we fit each
constrained linear class of Table~\ref{tab:structure} in closed form, under
one protocol. No neural network is trained.

\paragraph{Protocol.} ETTh1, ETTh2, ETTm1, ETTm2 with the standard
12/4/4-month split, and Electricity, Traffic, Weather and
Exchange with the standard 7:1:2 split; train-set standardization;
$L=720$, $H\in\{96,720\}$; all test windows; training
windows taken with stride 2 (hourly), 4 (15- and 10-min) and 1 (daily);
channel-independent pooling; window-mean removal as in SparseTSF.
Each estimator is ridge least squares with the ridge weight chosen on the
validation set from six values ($10^{-6}$ to $10^{-1}$,
relative to the mean diagonal of the Gram matrix). Widening this grid to
$10^{-8}$--$10^{2}$ changes no conclusion below. The three classes are: the full map $W$;
a low-pass input map, which reads only the first $k$ Fourier bins of the
input and maps them freely to the horizon (the input side of the FITS
constraint; FITS also band-limits its output, so its class is smaller); and
the phase-shared map $A\otimes I_w$ (the SparseTSF constraint). We sweep
$k$ and $w$. Code and results are released with the paper.

\begin{figure}[t]
\centering
\includegraphics[width=\linewidth]{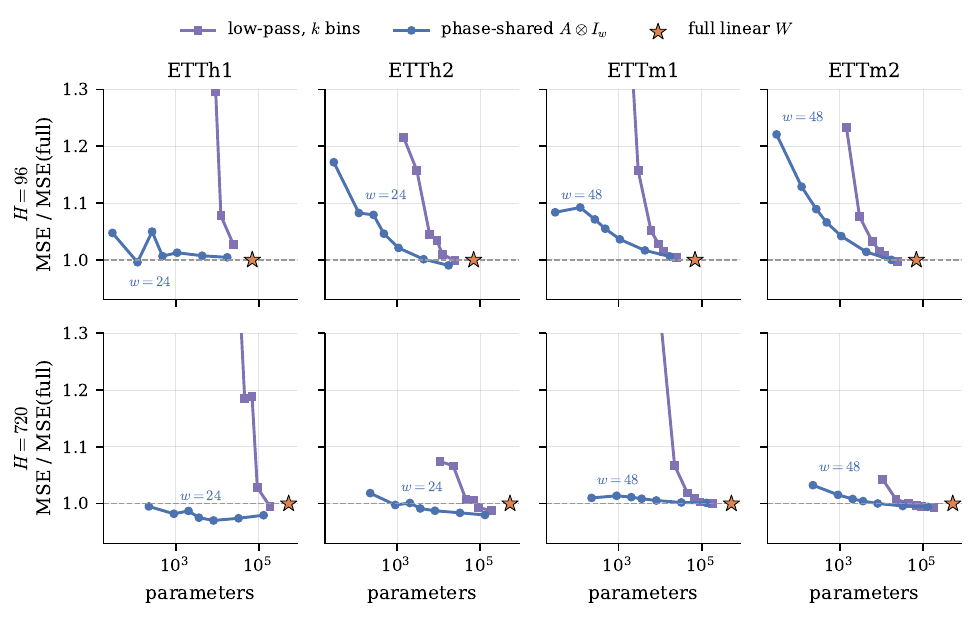}
\caption{Closed-form structured least squares under one protocol. Each curve sweeps the size of one constraint: input cutoff $k$ (low-pass input map) and period $w$ (phase-shared). Star and dotted line: full linear map. $L=720$.}
\label{fig:closedform}
\Description{Line plots of the test error of closed-form least-squares forecasters as the size of each constraint grows: the input cutoff for the low-pass input map and the period for the phase-shared map. A star and a dotted line mark the full linear map.}
\end{figure}

\paragraph{Findings.} Figure~\ref{fig:closedform} shows the results
on ETT, and Table~\ref{tab:naive} the main settings on all eight datasets.
\begin{enumerate}
\item \textbf{The Kronecker constraint is almost free on hourly data.} On
ETTh1 and ETTh2 at $H=720$, $w=24$ uses 900 parameters instead of 518{,}400
and matches or beats the full map (ETTh1: 0.426 vs.\ 0.434; ETTh2: 0.383
vs.\ 0.384). The same holds on the large hourly datasets:
on Electricity and Traffic at $H=720$, the 900 parameters stay within 1.5\%
of the full map (0.206 vs.\ 0.203; 0.451 vs.\ 0.449), and within 2\% (Traffic)
and 5\% (Electricity) at $H=96$ with 120 parameters.
\item \textbf{It also holds at a short horizon on ETTh1.} On ETTh1 ($H=96$),
the phase-shared map with $w=24$ matches the full map with 120 parameters
instead of 69{,}120 (0.371 vs.\ 0.373).
\item \textbf{It is not free everywhere.} On ETTh2 at $H=96$, and on ETTm1/ETTm2
at $H=96$ with $w=48$, the folded maps lose 8--22\% (e.g., ETTm2: 0.198
vs.\ 0.162). Smaller $w$ recovers the full-map error, which matches the
choice $w=4$ for ETTm in SparseTSF. Weather (10-min) behaves
the same way: folding at the daily period ($w=144$) loses 8.6\% at $H=720$
(0.348 vs.\ 0.320), and $w=48$ loses 27\% at $H=96$, while $w=2$ is within
0.5\% of the full map at both horizons.
\item \textbf{Low-pass input maps need the harmonics inside the band.} For hourly
data with $L=720$ the daily harmonics sit at bins 30, 60, \dots\ On ETTh1
($H=720$), cutoffs that exclude the first harmonic ($k\le16$) give 0.71;
including it ($k=32$) gives 0.51; including the second ($k=64$) gives 0.45;
$k=128$ reaches the full map (0.432 vs.\ 0.434). This is the logic of the
FITS input cutoff $k=(\lfloor L/T\rfloor+1)h+10$ with $h$ harmonics.
On Electricity and Traffic the same jumps appear at
$k=32$ and $k=64$, but even $k=128$ (four daily harmonics) stays 4--13\%
above the full map: their sharper daily profiles need more harmonics than
ETT. On Weather and Exchange, $k=128$ is within 1\% of the full map.
\item \textbf{No structure helps without a period.} On
Exchange, the negative control, folding loses more as $w$ grows (+25\% at
$w=24$, $H=96$), and no learned linear map beats the zero-parameter
last-value forecast (0.081 vs.\ 0.084 at $H=96$; 0.810 vs.\ 0.916 at
$H=720$).
\end{enumerate}
On the hourly benchmarks (ETTh1, ETTh2, Electricity,
Traffic), the headline claims of the lightweight literature
thus follow from the constraint itself, without a network or a training
trick. They also fail
where the folding assumption is weakest: short horizons,
high-resolution data (ETTm, Weather), where one period does
not dominate the window, and data without a period
(Exchange).

\paragraph{Classical reference.} A reviewer of any lightweight model should
ask how it compares with the zero-parameter \emph{seasonal naive} forecast,
$\hat y_{t+h}=x_{t+h-w\lceil h/w\rceil}$, the standard classical benchmark
\citep{hyndman2021fpp3}. Table~\ref{tab:naive} gives it under the same
protocol. The learned maps are 21--38\% better on every ETT
setting (e.g., ETTh1, $H=96$: 0.371 with 120 parameters vs.\ 0.512), so the
tiny models do learn more than the last cycle. On
Electricity, Traffic and Weather they are 21--68\% better. On Exchange, the
last-value forecast beats every learned map; with a wider ridge grid, the
validation set selects stronger regularization and the learned maps get
worse on test (full map at $H=720$: 1.10), a sign of the shift between the
validation and test periods of that series. Classical models (ETS, ARIMA) are rarely
reported on these benchmarks, and seasonal naive is almost never reported;
we recommend adding it to every table (Section~\ref{sec:protocol}).

\begin{table}[t]
\centering
\caption{Test MSE of the seasonal naive forecast (daily period, no
parameters; last value on Exchange), the phase-shared map
($w=24$ for hourly data; $w=48$, half a day, for 15-min data, since 96 does
not divide $L=720$; Weather: $w=48$ at $H=96$ and the daily
$w=144$ at $H=720$; Exchange: $w=24$, which matches no period; parameters in
parentheses) and the full linear map, all under the protocol
of Section~\ref{sec:closedform} (from code).}
\label{tab:naive}
\small
\begin{tabular}{lrccc}
\toprule
Dataset & $H$ & Seasonal naive & Phase-shared $A\otimes I_w$ & Full linear $W$\\
\midrule
ETTh1 & 96 & 0.512 & 0.371 (120) & 0.373 (69,120)\\
ETTh1 & 720 & 0.655 & 0.426 (900) & 0.434 (518,400)\\
ETTh2 & 96 & 0.391 & 0.293 (120) & 0.271 (69,120)\\
ETTh2 & 720 & 0.525 & 0.383 (900) & 0.384 (518,400)\\
ETTm1 & 96 & 0.423 & 0.333 (30) & 0.307 (69,120)\\
ETTm1 & 720 & 0.574 & 0.419 (225) & 0.415 (518,400)\\
ETTm2 & 96 & 0.263 & 0.198 (30) & 0.162 (69,120)\\
ETTm2 & 720 & 0.471 & 0.362 (225) & 0.351 (518,400)\\
\midrule
Electricity & 96 & 0.321 & 0.140 (120) & 0.133 (69,120)\\
Electricity & 720 & 0.367 & 0.206 (900) & 0.203 (518,400)\\
Traffic & 96 & 1.217 & 0.393 (120) & 0.385 (69,120)\\
Traffic & 720 & 1.185 & 0.451 (900) & 0.449 (518,400)\\
Weather & 96 & 0.317 & 0.214 (30) & 0.168 (69,120)\\
Weather & 720 & 0.443 & 0.348 (25) & 0.320 (518,400)\\
Exchange & 96 & 0.081 & 0.105 (120) & 0.084 (69,120)\\
Exchange & 720 & 0.810 & 0.968 (900) & 0.916 (518,400)\\
\bottomrule
\end{tabular}
\end{table}

\section{Practical Guidance}
\label{sec:practice}

\subsection{Deployment Considerations}
\label{sec:deploy}

Parameter count is the headline metric of this literature, but deployment
depends on latency, memory and energy. Energy per forecast is rarely
measured, so the size argument rests mostly on parameter counts and CPU
latency. We return to this gap in Section~\ref{sec:open}.

\paragraph{Latency.} On the single benchmark that reports CPU inference for
many models (Figure~\ref{fig:timeline}, right), latency follows size only
roughly. The linear-scale models (up to 1.04M parameters: TimeBase,
SparseTSF, FITS, DLinear) run in under 4\,ms per sample; the Transformers
need 30--250\,ms. Between Transformers, latency depends on
the attention pattern more than on size: PatchTST is the slowest (249\,ms)
with fewer parameters (8.69M) than Informer (22.45M, 73\,ms)
\citep[Table~3]{huang2025timebase}.

\paragraph{Compute versus size.} MACs and parameters can diverge. SSCNN has
about 5K parameters but more FLOPs than iTransformer because it keeps
full-length representations \citep{deng2024sscnn}; TimeMixer has 5.58M
parameters but 30.3G MACs, versus 1.79G for iTransformer with 5.47M
\citep[Table~3]{huang2025timebase}. Edge deployment should report MACs and
latency on the target device, not parameters alone.

\paragraph{Memory.} Training memory is dominated by activations. At batch
size 12 on Electricity, TimeBase needs 89\,MB and SparseTSF 125\,MB, versus
18\,GB for PatchTST \citep[Table~3]{huang2025timebase}. For
channel-dependent models, memory also grows with $C$ (e.g., TQNet's
attention over $C$ channel tokens).

\paragraph{Compression.} Generic compression (pruning, quantization,
distillation \citep{han2016deep,hinton2015distilling}) is almost unused in
this literature, which prefers designing small models directly. Whether a
pruned or quantized Transformer can match a folding model at equal size is
an open question.

\subsection{Which Model When}
\label{sec:guide}

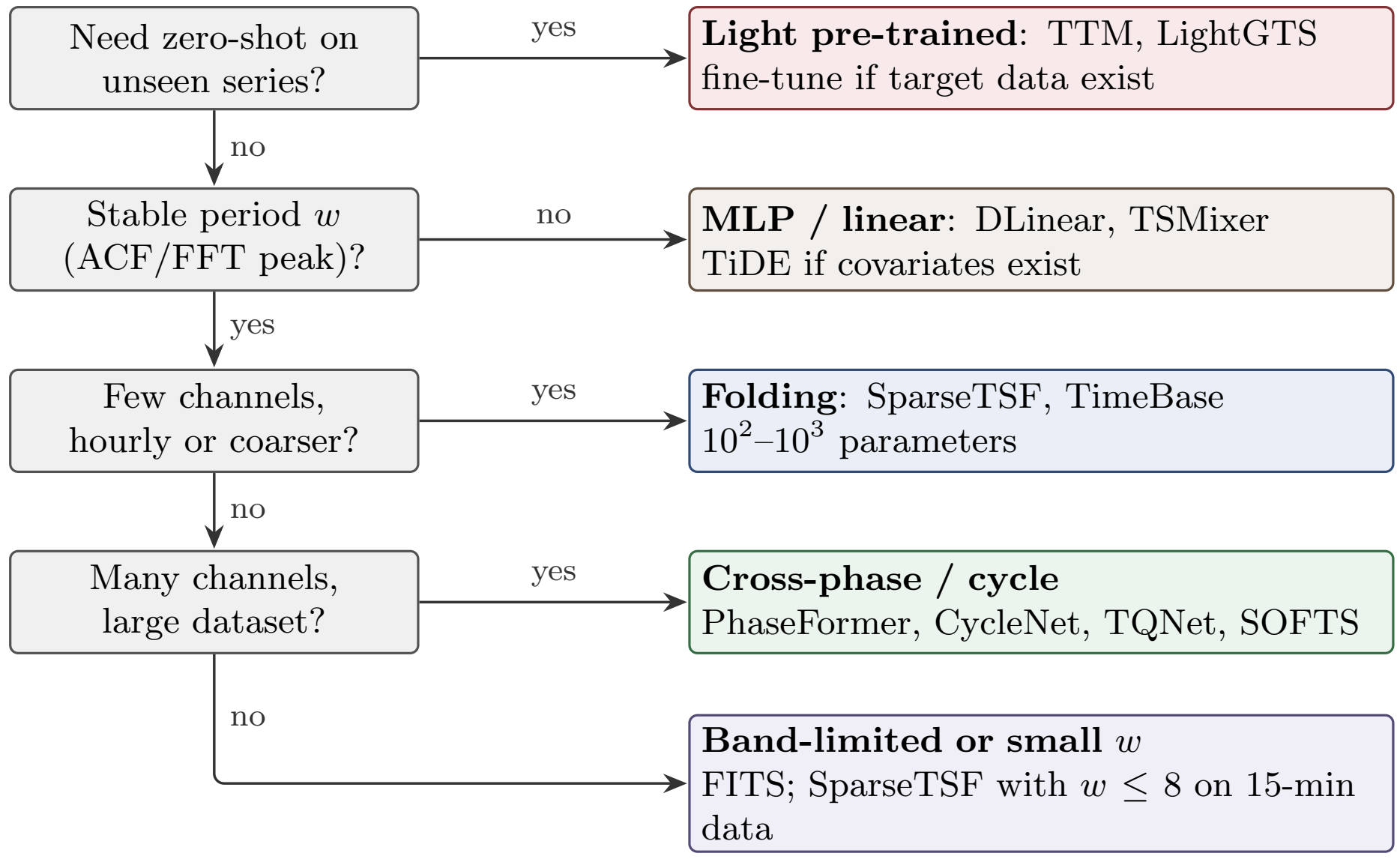
\begin{figure}[t]
\centering
\adjustbox{max width=0.95\linewidth}{\begin{tikzpicture}[x=1cm,y=1cm,
  q/.style={blk=gray,minimum width=36mm,minimum height=9mm},
  a/.style={blk=#1,minimum width=62mm,minimum height=9mm,align=left,text width=60mm}]
\node[q] (q1) at (0,0) {Need zero-shot on\\unseen series?};
\node[a=cAtt] (a1) at (7.3,0) {\textbf{Light pre-trained}: TTM, LightGTS\\fine-tune if target data exist};
\node[q] (q2) at (0,-1.6) {Stable period $w$\\(ACF/FFT peak)?};
\node[a=cPar] (a2) at (7.3,-1.6) {\textbf{MLP / linear}: DLinear, TSMixer\\TiDE if covariates exist};
\node[q] (q3) at (0,-3.2) {Few channels,\\hourly or coarser?};
\node[a=cIn] (a3) at (7.3,-3.2) {\textbf{Folding}: SparseTSF, TimeBase\\$10^2$--$10^3$ parameters};
\node[q] (q4) at (0,-4.8) {Many channels,\\large dataset?};
\node[a=cNorm] (a4) at (7.3,-4.8) {\textbf{Cross-phase / cycle}\\PhaseFormer, CycleNet, TQNet, SOFTS};
\node[a=cFreq] (a5) at (7.3,-6.4) {\textbf{Band-limited or small $w$}\\FITS; SparseTSF with $w\le8$ on 15-min data};
\draw[arr] (q1)--node[above,lbl]{yes}(a1);
\draw[arr] (q1)--node[right,lbl]{no}(q2);
\draw[arr] (q2)--node[above,lbl]{no}(a2);
\draw[arr] (q2)--node[right,lbl]{yes}(q3);
\draw[arr] (q3)--node[above,lbl]{yes}(a3);
\draw[arr] (q3)--node[right,lbl]{no}(q4);
\draw[arr] (q4)--node[above,lbl]{yes}(a4);
\draw[arr] (q4) |- node[pos=0.25,right,lbl]{no}(a5);
\end{tikzpicture}}
\caption{A model-selection guide derived from the evidence in this survey.}
\label{fig:guide}
\Description{Decision tree for choosing a model family. Questions in order: is zero-shot forecasting on unseen series needed (then light pre-trained models); is there a stable period, detected from the autocorrelation or spectrum peak (if not, MLP or linear models); are there few channels at hourly or coarser resolution (then period folding); are there many channels and a large dataset (then cross-phase or cycle models); otherwise band-limited or small-period models.}
\end{figure}

Figure~\ref{fig:guide} condenses the evidence of this survey into a
selection guide. It is a starting point, not a rule: every branch is
supported by results above, but data sets vary.
\begin{itemize}
\item \emph{Stable period, hourly or coarser, few channels:} folding models
reach full-map accuracy with $10^2$--$10^3$ parameters
(Section~\ref{sec:closedform}, Table~\ref{tab:acc}).
\item \emph{Many channels, large data:} add cross-phase or cross-channel
interaction (PhaseFormer, CycleNet, TQNet, SOFTS); the linear folding models
fall behind on Traffic and trail the interaction models
slightly on Electricity (Table~\ref{tab:acc}).
\item \emph{High-resolution data:} a daily fold costs accuracy at short
horizons (Section~\ref{sec:closedform}); use small $w$ or a band-limited map.
\item \emph{No stable period or with covariates:} use linear or MLP models
(DLinear, TSMixer, TiDE).
\item \emph{No training data:} use a small pre-trained model (TTM, LightGTS).
\end{itemize}

\subsection{Recommended Reporting Protocol}
\label{sec:protocol}

The audit suggests a short checklist. Each item addresses a discrepancy
documented above.
\begin{enumerate}
\item \textbf{Parameters from code.} Report the trainable parameter count
from the constructor of the released configuration, per dataset and
horizon, and state whether complex weights count once or twice and whether
per-channel weights are used.
\item \textbf{Same look-back, and a sweep.} Report results at a common
$L$ (e.g., 720) and at each model's best $L\in\{96,336,720\}$.
\item \textbf{Post-fix evaluation.} Use \texttt{drop\_last=False} at test
time and say so.
\item \textbf{Compute, not only size.} Report MACs, peak memory, and CPU
latency with stated hardware and batch size; SSCNN shows that size and
compute can diverge \citep{deng2024sscnn}.
\item \textbf{Classical reference.} Report the seasonal naive forecast;
it costs nothing and anchors the scale of every improvement
(Table~\ref{tab:naive}).
\item \textbf{Side information.} State whether the model uses the absolute
phase of the window (CycleNet, TQNet) or only the values.
\end{enumerate}

\section{Open Problems}
\label{sec:open}

\paragraph{Period selection.} Folding models need $w$. SparseTSF selects it from the
autocorrelation function and PhaseFormer from the dominant FFT component;
SparseTSF's ablation shows that the choice changes accuracy. Learning $w$, or folding at several
periods jointly, is open.

\paragraph{Multiple and drifting periods.} A single fold assumes one
dominant period. Daily and weekly cycles together (Traffic), or drifting
periods, break this assumption. A two-level fold, $w_1=24$ inside
$w_2=168$, gives a three-way tensor; predictors that share weights along
both periods are unexplored.

\paragraph{Channel dependence at low cost.} Most tiny models are CI. TQNet
adds channel attention but its size grows as $L^2$; SSCNN adds a spatial
component at higher compute. We are not aware of a channel-dependent
model below $10^4$ parameters that matches the folding models.

\paragraph{Aperiodic data.} On series without stable cycles, the structural
priors give little. PhaseFormer states this limitation explicitly
\citep{niu2026phaseformer}. A lightweight model needs a principled fallback
(e.g., to the unconstrained linear map) when no stable period is detected.

\paragraph{Theory.} The structured-linear view reduces most of these models
to constrained least squares on normalized windows. Generalization bounds
as a function of the constraint (Kronecker, band-limited), and
closed-form estimators for these classes, would turn many empirical
choices into analysis. Links to Koopman-operator forecasting
\citep{liu2023koopa} are a promising direction.

\paragraph{Energy and on-device evaluation.} Lightweight models are motivated
by constrained hardware, yet they are evaluated on server GPUs or desktop
CPUs, and energy per forecast is rarely measured. Benchmarks on
microcontrollers and single-board computers, with latency, memory and
energy reported together, would show whether parameter counts translate
into deployment gains.

\section{Conclusion}
\label{sec:conclusion}

Lightweight forecasters are best read as structured linear operators with
a light non-linearity from normalization. Frequency truncation gives a
band-limited map, period folding gives Kronecker structure, and explicit
cycles store periodicity as parameters. Our closed-form experiment shows
that, on hourly data (ETTh1, ETTh2, Electricity, Traffic)
at long horizons, the Kronecker constraint alone
keeps full-map accuracy to within 1.5\% with 576$\times$
fewer parameters, which explains why a few hundred parameters can be enough
on hourly data with a stable daily period. It costs more on
high-resolution data and helps nothing without a period (Exchange). On Traffic,
interaction across phases is needed; on Electricity it brings a smaller
gain. Our audit shows that the field's
main quantity, the parameter count, is reported inconsistently, and that
look-back choices can outweigh model differences. The models are also
useful: under one protocol they beat the seasonal naive forecast by
21--68\% on the seven periodic datasets, though not on
Exchange.
A common protocol would
make these results comparable.

\paragraph{Reproducibility.} The parameter-counting script and the
extracted tables are included with the source files of this paper and
will be released in a public repository.

\bibliographystyle{plainnat}
\bibliography{refs}

\appendix
\section{Counting Method and Numerical Checks}
\label{app:counting}

\paragraph{Parameter counting.} Each model is built with its repository's
own \texttt{Model.\_\_init\_\_}, unchanged, and its parameters are summed;
a complex parameter counts as two real parameters (PyTorch's
\texttt{numel} counts it once, which halves the complex layers of FITS and
Amplifier). Counts were first obtained by executing the constructors
against a minimal stand-in for \texttt{torch.nn} that records parameter
shapes, then re-checked by instantiating the same code in PyTorch~2: all
27 configurations tested agree exactly. They also reproduce every
published figure whose configuration we could identify
(Section~\ref{sec:audit}).

\paragraph{Configurations.} SparseTSF: $w=24$, linear variant. TimeBase:
$w=24$, $R=6$ (scripts for ETTh1). FITS: cutoff
$(\lfloor L/24\rfloor+1)\cdot2+10$ (repository default rule with $h=2$).
CycleNet and TQNet: $w=24$, $C=7$; TQNet $d=512$ (default).
PhaseFormer: $d=8$, one layer, 8 routers, positional embedding on.
Reference models: Time-Series-Library defaults.

\paragraph{Numerical checks.} With $L=H=720$, $w=24$ and random weights,
we verified that (i) the SparseTSF reshaping equals $(A\otimes I_w)\mathbf x$;
and (ii) the TimeBase forward pass acts on phase vectors with a map shared
across phases. Both checks passed to machine precision.

\paragraph{Released logs.} CycleNet and TQNet ship per-seed test results
(\texttt{result.txt}). Table~\ref{tab:lookback} averages them over seeds
and horizons.

\section{All Models at a Glance}
\label{app:master}

\begin{table}[h]
\centering
\caption{Master comparison of the models discussed. CI: channel-independent (?: not checked).
Period: side information the model needs ($w$ period, $t_0$ absolute phase, $T$ base period, $c$ cycle length).
Parameters: from official code at $L=720$, $H=96$, ETTh1 configuration, unless marked:
$^\ast$ released ETTh1 configuration at its own look-back; $^\ddagger$ small configuration ($d=8$);
$^\S$ reported by the paper, or by \citet{huang2025timebase} at $L=H=720$ on Electricity (SSCNN: ETT); n.c.\ not counted.}
\label{tab:master}
\small
\begin{tabular}{llllccr}
\toprule
Model & Venue & Family & Tokens & CI & Period & Params\\
\midrule
DLinear \citep{zeng2023dlinear} & AAAI'23 & Linear & point & \checkmark & -- & 138.4K\\
RLinear \citep{li2023rlinear} & arXiv'23 & Linear & point & \checkmark & -- & 69.2K\\
FITS \citep{xu2024fits} & ICLR'24 & Freq./basis & frequency & \checkmark & base $T$ & 11.8K\\
FreTS \citep{yi2023frets} & NeurIPS'23 & Freq./basis & frequency & -- & -- & 23.68M\\
FilterNet \citep{yi2024filternet} & NeurIPS'24 & Freq./basis & frequency & ? & -- & n.c.\\
Amplifier \citep{fei2025amplifier} & AAAI'25 & Freq./basis & frequency & \checkmark & -- & 55.1K$^\ast$\\
RAYQUAZA \citep{aityahia2026rayquaza} & OpenReview & Freq./basis & RBF atoms & ? & -- & n.c.\\
SparseTSF \citep{lin2024sparsetsf} & ICML'24 & Folding & phase & \checkmark & $w$ & 145\\
TimeBase \citep{huang2025timebase} & ICML'25 & Folding & phase & \checkmark & $w$ & 214\\
PhaseFormer \citep{niu2026phaseformer} & ICLR'26 & Folding & phase & \checkmark & $w$ & 1.7K$^\ddagger$\\
MoFo \citep{ma2025mofo} & NeurIPS'25 & Folding & 2D period & ? & $w$ & n.c.\\
PDF \citep{dai2024pdf} & ICLR'24 & Folding & 2D period & ? & FFT & n.c.\\
CycleNet \citep{lin2024cyclenet} & NeurIPS'24 & Cycle & point & \checkmark & $w$, $t_0$ & 69.4K\\
TQNet \citep{lin2025tqnet} & ICML'25 & Cycle & channel & -- & $w$, $t_0$ & 3.02M\\
SSCNN \citep{deng2024sscnn} & NeurIPS'24 & Cycle & point & -- & $c$ & $\sim$5K$^\S$\\
LightTS \citep{zhang2022lightts} & arXiv'22 & MLP & sampled & -- & -- & 182.8K\\
TSMixer \citep{chen2023tsmixer} & TMLR'23 & MLP & point & -- & -- & 1.56M\\
TiDE \citep{das2023tide} & TMLR'23 & MLP & point & \checkmark & -- & 4.51M\\
PITS \citep{lee2024pits} & ICLR'24 & MLP & patch & \checkmark & -- & 1.10M$^\ast$\\
xPatch \citep{stitsyuk2025xpatch} & AAAI'25 & MLP/CNN & patch & \checkmark & -- & 144.0K$^\ast$\\
SOFTS \citep{han2024softs} & NeurIPS'24 & MLP & channel & -- & -- & 510.8K$^\ast$\\
TimeMixer \citep{wang2024timemixer} & ICLR'24 & MLP & multi-scale & \checkmark & -- & 5.58M$^\S$\\
LiteMixer \citep{aityahia2025litemixer} & WINCOM'25 & MLP & multi-scale & ? & -- & n.c.\\
S-Mamba \citep{wang2024smamba} & arXiv'24 & SSM & variate & -- & -- & 1.15M$^\ast$\\
N-BEATS \citep{oreshkin2020nbeats} & ICLR'20 & MLP & point & \checkmark & -- & n.c.\\
SegRNN \citep{lin2023segrnn} & arXiv'23 & RNN & segment & \checkmark & -- & 1.63M\\
TimeKAN \citep{huang2025timekan} & ICLR'25 & KAN & multi-scale & \checkmark & -- & 12.8K$^\ast$\\
TTM \citep{ekambaram2024ttm} & NeurIPS'24 & Pre-trained & patch & \checkmark & -- & $\ge$1M$^\S$\\
LightGTS \citep{wang2025lightgts} & ICML'25 & Pre-trained & period patch & \checkmark & $w$ & 1M/4M$^\S$\\
Chronos \citep{ansari2024chronos} & TMLR'24 & Pre-trained & quantized & \checkmark & -- & 20M--710M$^\S$\\
Moirai \citep{woo2024moirai} & ICML'24 & Pre-trained & patch & -- & -- & 14M--311M$^\S$\\
TimesFM \citep{das2024timesfm} & ICML'24 & Pre-trained & patch & \checkmark & -- & 200M$^\S$\\
PatchTST \citep{nie2023patchtst} & ICLR'23 & Transformer & patch & \checkmark & -- & 10.74M\\
iTransformer \citep{liu2024itransformer} & ICLR'24 & Transformer & variate & -- & -- & 6.72M\\
\bottomrule
\end{tabular}
\end{table}

\section{Official Code Used in This Survey}
\label{app:code}

Table~\ref{tab:code} lists the repositories we cloned for the audit
(checked in September 2026) and the commits at which the parameter counts and
test-loader checks were re-verified (October 2026). Parameter counts in this
paper are computed from these implementations. The accompanying code, which will be released publicly, contains a single
benchmark built on the Time-Series-Library that includes these
implementations unchanged at pinned commits, the official training scripts
translated to one command-line interface, and the counting and profiling
tools used here.

\begin{table}[h]
\centering
\caption{Official repositories (all on GitHub).}
\label{tab:code}
\small
\begin{tabular}{lll}
\toprule
Model & Repository & Commit\\
\midrule
DLinear, NLinear & \texttt{cure-lab/LTSF-Linear} & \texttt{0c11366}\\
RLinear & \texttt{plumprc/RTSF} & \texttt{0fec001}\\
FITS & \texttt{VEWOXIC/FITS} & \texttt{d040bb0}\\
FreTS & \texttt{aikunyi/FreTS} & \texttt{6de28ab}\\
Amplifier & \texttt{aikunyi/Amplifier} & \texttt{6cc0893}\\
SparseTSF & \texttt{lss-1138/SparseTSF} & \texttt{b8c2740}\\
SegRNN & \texttt{lss-1138/SegRNN} & \texttt{8e869ec}\\
CycleNet & \texttt{ACAT-SCUT/CycleNet} & \texttt{d807e51}\\
TQNet & \texttt{ACAT-SCUT/TQNet} & \texttt{15e19cb}\\
TimeBase & \texttt{hqh0728/TimeBase} & \texttt{369b330}\\
PhaseFormer & \texttt{neumyor/PhaseFormer\_TSL} & \texttt{ed1db61}\\
TimeKAN & \texttt{huangst21/TimeKAN} & \texttt{3a7c366}\\
SOFTS & \texttt{Secilia-Cxy/SOFTS} & \texttt{f5d35fd}\\
xPatch & \texttt{stitsyuk/xPatch} & \texttt{d12eeca}\\
PITS & \texttt{seunghan96/pits} & \texttt{a4c56af}\\
S-Mamba & \texttt{wzhwzhwzh0921/S-D-Mamba} & \texttt{e7e8bf0}\\
Reference models & \texttt{thuml/Time-Series-Library} & \texttt{4e938a1}\\
\bottomrule
\end{tabular}
\end{table}

\paragraph{Reproducing the tables and figures.} Table~\ref{tab:repro} maps
each quantitative table and figure to the script that produces it. Scripts
are in the \texttt{survey/} folder of the accompanying benchmark.
Values transcribed from published papers are marked as such.

\begin{table}[h]
\centering
\caption{Script behind each quantitative table and figure (folder
\texttt{survey/} of the accompanying benchmark).}
\label{tab:repro}
\small
\begin{tabular}{@{}>{\raggedright\arraybackslash}p{4.2cm}>{\raggedright\arraybackslash}p{8.4cm}@{}}
\toprule
Table or figure & Source\\
\midrule
Tables~\ref{tab:count} and~\ref{tab:structure}, Figure~\ref{fig:sweep} & \texttt{audit/sweep.py}, \texttt{new\_counts.py}, \texttt{more2.py}, re-checked in PyTorch by \texttt{tests/test\_models.py}\\
Table~\ref{tab:inconsistent} & reported values transcribed from the cited tables; ``From code'' from \texttt{audit/sweep.py}\\
Table~\ref{tab:droplast} & \texttt{audit/check\_test\_loader.py} on the commits of Table~\ref{tab:code}\\
Table~\ref{tab:acc}, Figure~\ref{fig:paretolit} & accuracy transcribed from \citet[Table~1]{niu2026phaseformer}, sizes from \texttt{audit/sweep.py}, plot by \texttt{experiments/plots2.py}\\
Table~\ref{tab:lookback}, Figure~\ref{fig:lookback} & CycleNet rows: \texttt{audit/lookback\_from\_logs.py} on the released \texttt{result.txt}. SparseTSF rows: \citet[Table~12]{lin2024sparsetsf}\\
Table~\ref{tab:naive}, Figure~\ref{fig:closedform} & \texttt{experiments/closed\_form.py}, \texttt{naive.py}, \texttt{plots2.py}; Electricity, Traffic, Weather and Exchange rows: \texttt{experiments/closed\_form\_ext.py}\\
Figure~\ref{fig:timeline} & values transcribed from \citet[Table~3]{huang2025timebase}, plot by \texttt{experiments/plots3.py}\\
\bottomrule
\end{tabular}
\end{table}

\end{document}